\pdfoutput=1
\documentclass[10pt]{article}
\usepackage[utf8]{inputenc}
\usepackage[T1]{fontenc}
\usepackage{arxiv}
\usepackage{amsmath,amssymb,amsfonts,amsthm}
\usepackage{graphicx}
\usepackage{textcomp}
\usepackage[table]{xcolor}
\usepackage{booktabs}
\usepackage{array}
\usepackage{multirow}
\usepackage{adjustbox}
\usepackage{threeparttable}
\usepackage{placeins}
\usepackage{tikz}
\usetikzlibrary{arrows.meta,positioning,calc}
\usepackage{algorithm}
\usepackage{algpseudocode}
\usepackage{enumitem}
\usepackage{microtype}
\usepackage{cite}
\usepackage{url}
\usepackage[colorlinks=true,linkcolor=blue!50!black,citecolor=blue!50!black,urlcolor=blue!50!black]{hyperref}
\theoremstyle{plain}
\newtheorem{proposition}{Proposition}
\newtheorem{corollary}{Corollary}
\newcommand{\trunc}{\operatorname{trunc}}
\newcommand{\pmstd}[2]{#1{\scriptsize$\pm$#2}}
\setlist[itemize]{leftmargin=1.4em,itemsep=2pt,topsep=3pt}

\title{Sensitivity-Gated Accumulator Caching for\\On-Device Learning of Low-Bit Hand Networks}
\renewcommand{\shorttitle}{Sensitivity-Gated Accumulator Caching}
\author{%
Soroor Ghandali\thanks{Corresponding author: \texttt{soroor.ghandali@sjsu.edu}.}\qquad Kanika Mamgain\qquad Manjot Kaur\qquad Akanksha Shukla\qquad Shashira Guntuka\\[5pt]
Department of Applied Data Science, San Jos\'e State University, San Jos\'e, CA, USA\\[2pt]
{\footnotesize\texttt{\{soroor.ghandali, kanika.mamgain, manjot.kaur02, akanksha.shukla, shashira.guntuka\}@sjsu.edu}}}

\begin{document}

\maketitle

\begin{abstract}
Small hand-tracking networks are deployed as 4- and 8-bit codes, and they should keep adapting to a user's camera after deployment, on the device and within roughly the memory of the model itself. Quantization-aware training cannot do this cheaply, because it keeps a 32-bit master copy of every weight together with its optimizer state, about 96 bits per weight. Training the codes directly with stochastic rounding needs no copy but injects rounding noise at every step, and the methods that remove this noise keep a residual buffer for every weight. We introduce sensitivity-gated accumulator caching (SGAC), which keeps residual state for only a small and changing set of weight groups. A set-associative cache of 8-bit accumulators collects the sub-step updates of the groups with the largest Fisher-weighted squared code step and writes them into the codes as whole steps; an evicted line is flushed by stochastic rounding, and a group without a line does not move until it is admitted. We prove that the write-back is unbiased and show, under a local quadratic model, that rounding a sub-step update costs loss of the same order as the update's own gain, so that rounding pays only where the expected gradient is large relative to the curvature. On a 383k-parameter hand-keypoint network, SGAC trains with 8.1 bits of state per weight. Scored on the 4/8-bit network that the device runs after adaptation, it has the highest mean accuracy of nine deployed low-bit networks on two simulated cameras. On twelve ImageNet-C corruptions it has the highest average of corrupted and clean accuracy of seven deployed methods, ahead of the PEQA-style, ECO-style and direct low-bit baselines with confidence intervals that exclude zero, and it matches an 8-bit accumulator kept for every weight with 56\% of that state.
\end{abstract}

\keywords{On-device learning \and low-bit training \and quantization \and stochastic rounding \and domain adaptation \and hand pose estimation}

\section{Introduction}

\begin{figure}[!tbp]
\centering
\resizebox{0.93\textwidth}{!}{
\definecolor{bF}{HTML}{DAE8FC}\definecolor{bD}{HTML}{6C8EBF}\definecolor{bP}{HTML}{F2F7FE}
\definecolor{gF}{HTML}{D5E8D4}\definecolor{gD}{HTML}{82B366}\definecolor{gP}{HTML}{F3F9F2}
\definecolor{pF}{HTML}{E1D5E7}\definecolor{pD}{HTML}{9673A6}\definecolor{pP}{HTML}{F7F2F9}
\definecolor{yF}{HTML}{FFF2CC}\definecolor{yD}{HTML}{D6B656}
\definecolor{oF}{HTML}{FFE6CC}\definecolor{oD}{HTML}{D79B00}\definecolor{oP}{HTML}{FFF8F0}
\definecolor{nF}{HTML}{F5F5F5}\definecolor{nD}{HTML}{666666}
\begin{tikzpicture}[font=\footnotesize, >=Stealth,
  blk/.style={draw, rounded corners=2pt, align=center, inner sep=2pt, line width=0.6pt},
  panel/.style={rounded corners=5pt, line width=0.6pt},
  ttl/.style={font=\scriptsize\bfseries, anchor=west},
  arr/.style={->, line width=0.6pt, draw=black!75},
  lbl/.style={font=\scriptsize}]
\fill[bP, draw=bD, panel] (1.35,-2.05) rectangle (6.75,2.25);
\fill[gP, draw=gD, panel] (7.55,-2.05) rectangle (10.25,2.25);
\fill[pP, draw=pD, panel] (10.75,-2.05) rectangle (13.35,3.60);
\fill[oP, draw=oD, panel] (1.35,-4.70) rectangle (12.95,-2.60);
\node[ttl] at (1.45,2.00) {Perception \textmd{(App.~\ref{app:system})}};
\node[ttl] at (7.65,2.00) {Gestures \textmd{(App.~\ref{app:system})}};
\node[ttl] at (10.85,3.35) {Intent \textmd{(App.~\ref{app:system})}};
\node[font=\scriptsize, align=left, anchor=west] at (1.50,-3.65) {\textbf{Learning in}\\\textbf{low-bit memory:}\\\textbf{SGAC} (\S\ref{sec:method},\\Algorithm~\ref{alg:sgac})};
\node[blk, fill=nF, draw=nD, minimum width=16mm, minimum height=10mm] (cam) at (0,0) {Webcam\\frame $I_t$};
\node[blk, fill=nF, draw=nD, minimum width=17mm, minimum height=8mm] (p) at (14.55,1.15) {person\\(right hand)};
\node[blk, fill=nF, draw=nD, minimum width=17mm, minimum height=8mm] (c) at (14.55,-1.15) {cat\\(left hand)};
\node[blk, fill=bF, draw=bD, minimum width=18mm, minimum height=10mm] (d) at (2.55,0) {PalmNet-D\\{\scriptsize hands + face, 157k}};
\node[blk, fill=bF, draw=bD, minimum width=19mm, minimum height=8mm] (kr) at (5.45,1.15) {PalmNet-K\\{\scriptsize 383k, 4/8-bit}};
\node[blk, fill=bF, draw=bD, minimum width=19mm, minimum height=8mm] (kl) at (5.45,-1.15) {PalmNet-K\\{\scriptsize 383k, 4/8-bit}};
\node[blk, fill=yF, draw=yD, minimum width=19mm, minimum height=8.5mm, font=\scriptsize] (gate) at (5.45,0) {staggered motion\\gate + buffer};
\node[blk, fill=gF, draw=gD, minimum width=22mm, minimum height=8mm] (tr) at (8.90,1.15) {GestureTCN, 47k\\+ FDR gate};
\node[blk, fill=gF, draw=gD, minimum width=22mm, minimum height=8mm] (tl) at (8.90,-1.15) {GestureTCN, 47k\\+ FDR gate};
\node[blk, fill=yF, draw=yD, minimum width=22mm, minimum height=8.5mm, font=\scriptsize] (mem) at (8.90,0) {gesture memory\\(8/4-bit keys)};
\node[blk, fill=pF, draw=pD, minimum width=21mm, minimum height=31mm] (g) at (12.05,0) {Intent\\grammar\\[2pt]$\mathcal{G}(\cdot,s_t)$};
\node[blk, fill=pF, draw=pD, minimum width=21mm, minimum height=7.5mm] (snd) at (12.05,2.70) {SoundTag, 218k\\(microphone)};
\node[blk, fill=bF, draw=bD, minimum width=26mm, minimum height=10mm, inner xsep=5pt] (rom) at (5.45,-3.65) {codes (``ROM''): 4/8-bit $q$\\scales $s$, offsets $\beta$};
\node[blk, fill=oF, draw=oD, minimum width=26mm, minimum height=10mm, inner xsep=5pt] (sram) at (10.75,-3.65) {cache (``SRAM''): 23\% of\\groups, 8-bit accum.\ $a$};
\draw[arr] (cam) -- (d);
\draw[arr] (d.east) -- (3.95,0) |- (kr.west);
\draw[arr] (d.east) -- (3.95,0) |- (kl.west);
\draw[arr] (gate.north) -- (kr.south);
\draw[arr] (gate.south) -- (kl.north);
\draw[arr] (kr.east) -- node[lbl, above]{$P^R_t$} (tr.west);
\draw[arr] (kl.east) -- node[lbl, below]{$P^L_t$} (tl.west);
\draw[arr] (mem.north) -- (tr.south);
\draw[arr] (mem.south) -- (tl.north);
\draw[arr] (tr.east) -- node[lbl, above]{$y^R_t$} (tr.east -| g.west);
\draw[arr] (tl.east) -- node[lbl, below]{$y^L_t$} (tl.east -| g.west);
\draw[arr] (snd.south) -- (g.north);
\draw[arr] (g.east |- p.west) -- (p.west);
\draw[arr] (g.east |- c.west) -- (c.west);
\draw[arr] ([yshift=2.5mm]rom.east) -- node[lbl, above]{admit by $\pi_g$} ([yshift=2.5mm]sram.west);
\draw[arr] ([yshift=-2.5mm]sram.west) -- node[lbl, below]{commit / flush} ([yshift=-2.5mm]rom.west -| rom.east);
\draw[->, line width=0.9pt, draw=oD, dashed] (rom.north) -- node[lbl, right, text=black]{updated in place} (kl.south);
\end{tikzpicture}}
\caption{Overview of the two-hand system (top) and of SGAC (bottom); each panel is described in the section named in its title. For each hand, PalmNet-K turns the frame $I_t$ into keypoints $P_t^R$ or $P_t^L$, GestureTCN turns them into gesture tokens $y_t^R$ or $y_t^L$, and the intent grammar $\mathcal{G}$ turns tokens into actions given the scene state $s_t$. SGAC updates the deployed 4/8-bit codes in place through a small cache of accumulators.}
\label{fig:system}
\end{figure}
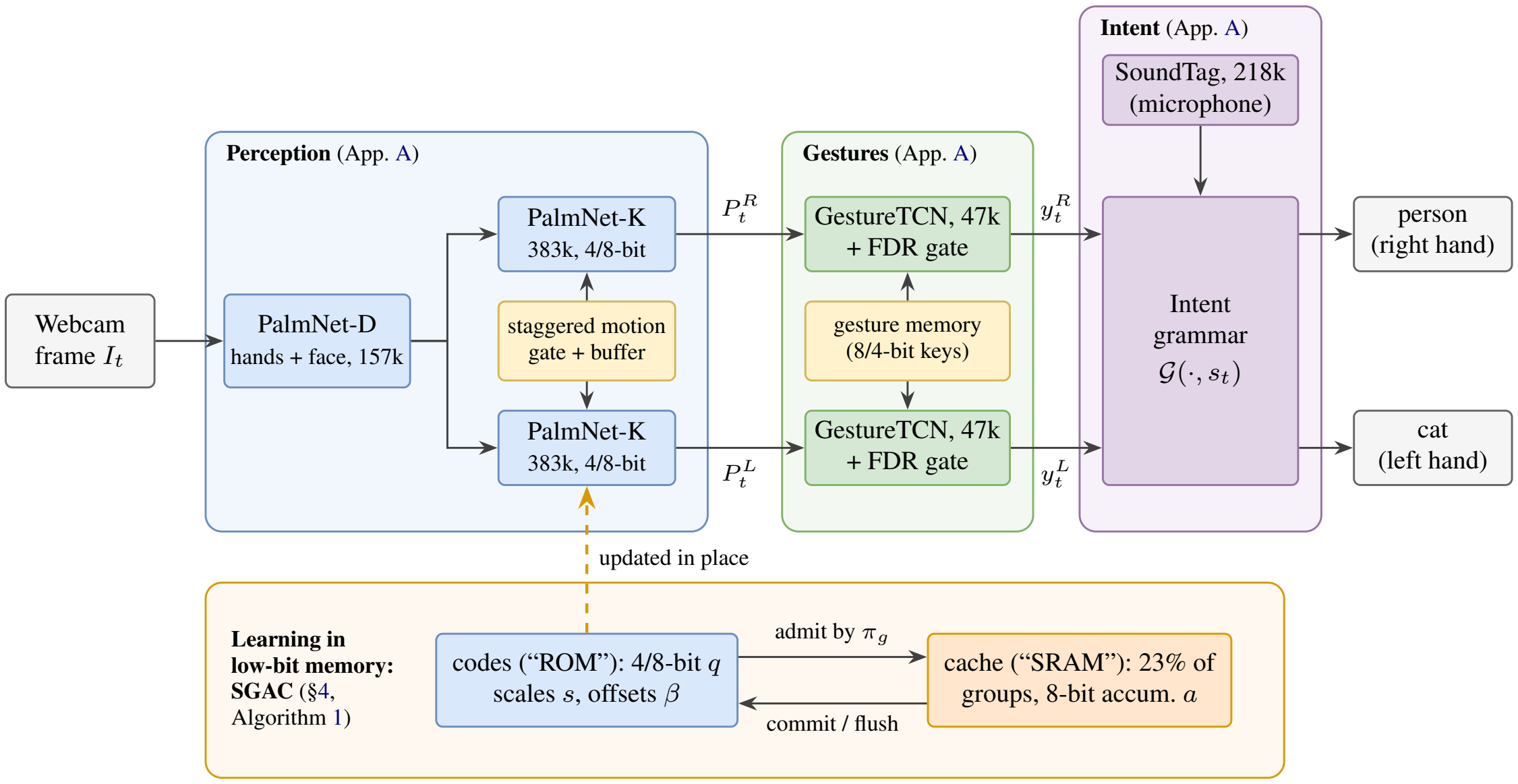

Hand interfaces that run on a laptop or in a browser rely on small networks. The webcam has to be read in a few tens of milliseconds per frame, often for two hands at once, and the models are therefore stored compactly, typically as 4- or 8-bit integer codes with a scale and an offset per group of weights \cite{jacob2018,lsq}. Keypoint networks trained on public datasets, however, lose accuracy when the camera, the lighting or the user changes \cite{freihand,hamer}. A natural remedy is to keep adapting the network on the device, where the user's images already are \cite{tinytl,ondevice256}. The network that has to adapt is then the low-bit network itself.

This is harder than it sounds, because the updates of an adaptation are tiny compared with the grid on which a low-bit weight lives. In our keypoint network the median code step of a 4-bit group is 0.011, while a typical step of an adaptive optimizer changes a weight by about $3\times10^{-4}$, roughly 3\% of one code step. A code can only move by whole steps, so such an update cannot be written as it is. Quantization-aware training solves the problem by keeping a 32-bit master copy of every weight with two Adam moments \cite{adam,jacob2018,lsq}; for our 383k-weight network that is 4.6\,MB of training state next to a 0.23\,MB model. Adapter methods such as QLoRA \cite{lora,qlora} avoid the copy but leave the codes frozen, so the adapters must either stay beside the codes at inference or be merged into them, where every change below half a step is rounded away. Stochastic rounding \cite{gupta2015,smgd,dqt} moves the codes directly and needs no copy, yet it adds noise to every weight at every step \cite{li2017}. Methods that remove this noise keep a residual for every weight: an accumulator in in-memory training \cite{nandakumar2020}, a gradient average in Bop \cite{bop}, a low-bit buffer in Error Diffusion \cite{errordiffusion}, or a momentum that absorbs the rounding error in ECO \cite{eco}.

Our starting point is the observation that not every weight needs its residual all the time. During a short adaptation, most of the useful change happens in a minority of sensitive weight groups. For the other groups, rounding can do more harm than good. Once the network has partly adapted, the new-camera update begins to work against the images it already knows, and rounding adds noise on top; in our ablation, letting these groups wait instead of rounding them is worth 1.4 points, most of it on the original images. Sensitivity-gated accumulator caching (SGAC) puts this into practice with a small cache, organized like a processor cache, that holds 8-bit sub-step accumulators for the groups with the largest Fisher-weighted squared code step. A cached group commits its accumulated update to the codes in whole steps. When its line is evicted, the remainder is flushed by stochastic rounding, so nothing is lost in expectation, and every group without a line waits. We call this policy \emph{wait-or-write-back}. Since the codes serve as the only copy of the convolution weights, the network on the device is the trained network after every step, and no conversion follows the adaptation.

The paper makes three contributions.
\begin{itemize}
\item \textbf{A training rule} (Section~\ref{sec:method}) that allocates residual state to a changing, sensitivity-ranked subset of weight groups and lets the rest wait. Table~\ref{tab:contrast} in Section~\ref{sec:related} places it among existing methods; to our knowledge, no prior method combines group-level, reassignable residual state with deferred rather than rounded updates for the weights it does not cover.
\item \textbf{An analysis} (Section~\ref{sec:analysis}) showing that the write-back is unbiased, that stochastic rounding of a sub-step update costs loss at first order in the step, and when waiting is better than rounding for an objective that weighs the new and the original data. Measured with unbiased Hessian estimates on our network, the share of weight groups for which waiting is better grows from 6\% at the start of adaptation to 37\% after 400 steps, as the update turns against the original images.
\item \textbf{A deployed-format evaluation} (Section~\ref{sec:exp}) that scores every method on the 4/8-bit network the device runs after adaptation, with step sizes chosen on validation data. Under this protocol, adapters that look strong while kept in full precision lose up to 5.8 points on the new camera once they are merged, and SGAC's deployed network has the highest mean accuracy of the nine deployed low-bit networks we compare on two simulated cameras, with 8.1 bits of training state per weight and 2.3 to 29 times less state traffic per step than its full-precision and adapter counterparts.
\end{itemize}

We developed SGAC for a two-hand webcam interface in which the right hand directs one character and the left hand another (Fig.~\ref{fig:system}); every network in it is trained from scratch and stored at 4/8 bits. Section~\ref{sec:system} summarizes this system, and Appendices~\ref{app:system} and~\ref{app:results} describe its other components and their results. None of that material is needed to follow or reproduce the method.

\section{Related Work}\label{sec:related}

\textbf{Low-bit training without a master copy.} Training directly on low-bit weights with stochastic rounding (SR) goes back to Gupta et al.~\cite{gupta2015}, and later work analyzed it \cite{li2017,smgd} and scaled it to language models \cite{dqt}. Li et al.~\cite{li2017} showed that SR loses the greedy, exploitative behavior of full-precision training near a minimum, which is consistent with our finding that rounding hurts most after the network has adapted. Several methods add a residual to remove the rounding noise. In-memory mixed-precision training keeps a high-precision accumulator for every weight and transfers whole device steps when it crosses a threshold \cite{nandakumar2020}. Bop keeps a real-valued gradient average per binary weight and flips the weight when the average passes a threshold \cite{bop}. Error Diffusion replaces the latent weight of a binary network with a 2--8-bit residual per weight \cite{errordiffusion}, and ECO folds the rounding error into the optimizer momentum \cite{eco}. All of these keep residual state either for every weight or for none.

\textbf{Adapters, sparse and grid-only fine-tuning.} LoRA and QLoRA train low-rank side modules over a frozen base \cite{lora,qlora}; QA-LoRA restricts them so that the merge into group-wise offsets is exact \cite{qalora}, and LoQT merges a low-rank factor into the quantized weights at intervals \cite{loqt}. SpIEL re-selects a sparse set of full-precision deltas during fine-tuning over a 4-bit base \cite{spiel}. PEQA tunes only the quantization scales \cite{peqa}, and BitFit and TinyTL update only biases \cite{bitfit,tinytl}. FISH Mask fixes a sparse set of parameters by their Fisher information \cite{fishmask}, sparse on-device training updates layers or channels chosen offline \cite{ondevice256}, and a recent study protects a gradient-ranked budget of weights with full-precision copies \cite{shang2026}. These methods change the network beside the codes, or only its grid; when a change has to reach the codes, it is rounded at merge time.

\textbf{Sensitivity scores and caches.} Fisher information or Hessian traces multiplied by the squared quantization step are used to choose bit widths after or during training \cite{fit,qbitopt}; SGAC uses the same product to decide which groups receive accumulator precision while the network learns. Outlier weights are commonly kept at higher precision \cite{spqr}. A set-associative cache of full-precision embedding rows over a low-precision table has been proposed for recommendation models, with rows chosen by access frequency \cite{mpecache}. Compute-in-memory designs store frozen weights in ROM and train a small SRAM part such as a residual branch \cite{cimromsram,cimromsram2}; SGAC instead trains the stored codes themselves through the cache.

\textbf{Re-parameterized convolutions.} Structural re-parameterization trains several convolution branches and merges them into one kernel for inference \cite{acnet,dbb}, and the same effect can be produced by reshaping the gradient of the merged kernel \cite{repopt,sgs}. Our compass gradient (Section~\ref{sec:method}) applies this equivalence to the six-branch block of our networks, so that the folded kernel on the device can still be trained.

\begin{table}[!tbp]
\caption{How each method handles an update smaller than one code step.}
\label{tab:contrast}
\centering
\footnotesize
\setlength{\tabcolsep}{3pt}\renewcommand{\arraystretch}{1.12}
\begin{adjustbox}{max width=\textwidth}
\begin{tabular}{@{}>{\raggedright\arraybackslash}p{3.3cm}>{\raggedright\arraybackslash}p{2.5cm}>{\raggedright\arraybackslash}p{3.0cm}>{\raggedright\arraybackslash}p{2.2cm}>{\raggedright\arraybackslash}p{4.6cm}@{}}
\toprule
Method & Residual kept for & Residual state & Weights without residual & How an update reaches the deployed codes \\[1pt]
\midrule
QAT, master copy \cite{jacob2018,lsq} & every weight & 32-bit copy + Adam & -- & codes re-made from the copy at every step \\
Direct SR \cite{gupta2015,smgd,dqt} & none & -- & rounded at every step & every step, with rounding noise \\
In-memory mixed precision \cite{nandakumar2020} & every weight & 32-bit accumulator & -- & whole device steps transferred when the accumulator passes a threshold \\
Bop \cite{bop} & every weight & real-valued gradient average & -- & sign flip past a threshold (binary weights) \\
Error Diffusion \cite{errordiffusion} & every weight & 2--8-bit residual, SR-quantized & -- & flip when the residual crosses a grid level \\
ECO \cite{eco} & every weight & momentum, error fed back & -- & every step, rounding error injected into momentum \\
FISH Mask \cite{fishmask} & fixed Fisher subset & full-precision copy & frozen & -- (full-precision training) \\
SpIEL \cite{spiel} & changing sparse subset & fp deltas + Adam & frozen & kept beside the codes, or rounded at merge \\
QLoRA \cite{qlora} / LoQT \cite{loqt} & none (codes frozen) & low-rank adapters + Adam & frozen & merged and rounded, at the end or periodically \\
QA-LoRA \cite{qalora} / PEQA \cite{peqa} & none & group adapters / scales + Adam & frozen & exact, but only offsets or scales change \\
\textbf{SGAC (ours)} & \textbf{changing, $\pi_g$-ranked 23\% of groups} & \textbf{8-bit sub-step accumulator} & \textbf{wait (codes unchanged)} & \textbf{whole steps on commit; unbiased flush on eviction} \\
\bottomrule
\end{tabular}
\end{adjustbox}
\end{table}

\textbf{What SGAC adds.} Table~\ref{tab:contrast} sorts the closest methods by where they keep the residual of a sub-step update and by what happens to a weight that has none. SGAC differs in three respects. First, its residual state is allocated per group, by sensitivity, and reassigned during training, rather than kept for every weight. Second, a group without state waits instead of being rounded; in SR, ECO and Error Diffusion every weight moves at every step. Third, an evicted line is flushed by stochastic rounding, so reassigning the cache never biases the update (Proposition~\ref{prop:unbiased}). The ablation in Section~\ref{sec:ablation} measures the first two: with the same codes and the same update, an 8-bit accumulator for every weight, the arrangement of in-memory training, reaches the same accuracy as SGAC with 14.6 instead of 8.1 bits of state; a 2-bit residual for every weight, the arrangement of Error Diffusion at roughly the same memory (8.6 bits), is 1.6 points less accurate; and rounding the uncached groups instead of letting them wait costs 1.4 points.

\section{Problem Formulation}
\label{sec:problem}

\textbf{Notation.} The weights of a layer $l$ are cut into groups of $n_g=64$, and the layer has a code width of $b_l$ bits. Group $g$ stores integer codes $q_i\in\{0,\dots,2^{b_l}-1\}$, one code step (scale) $s_g$ and one offset $\beta_g$, and a weight is $W_i=s_g\,q_i+\beta_g$. An update $\Delta W_i$ is measured in code steps as $d_i=\Delta W_i/s_g$; most $|d_i|$ are far below one. $\mathcal{L}$ is the training loss, $g_i=\partial\mathcal{L}/\partial W_i$ the gradient of one batch, and $F_g$ a running average of $g_i^2$ over the group, an estimate of its Fisher information.

\textbf{Learning in low-bit memory.} After deployment, a network is stored as codes $q$ with per-group scales $s$ and offsets $\beta$. When the camera, the lighting or the user changes, it meets a new distribution $\mathcal{D}'$ and should adapt on the device:
\begin{equation}
\begin{aligned}
\min_{q}\ \ &\mathbb{E}_{(x,y)\sim\mathcal{D}'}\,\mathcal{L}\big(f(x;W(q,s,\beta)),y\big)\\
\text{s.t.}\ \ &q_i\in\{0,\dots,2^{b_l}-1\},\qquad M\le \bar b+B,
\end{aligned}
\label{eq:p3}
\end{equation}
where $M$ is the persistent training state per weight, $\bar b$ is the average code width and $B$ is a small budget of extra bits. As in supervised domain adaptation, labeled samples from $\mathcal{D}'$ are available; in our system they come from a two-minute guided calibration (Appendix~\ref{app:system}). The network must remain deployable after every step, and it should keep its accuracy on the original distribution $\mathcal{D}$. Success is therefore measured on the stored network $W(q,s,\beta)$ after adaptation: a method that learns in another format, such as full-precision weights or adapters, has to be converted back into codes before the device can run it.

\section{Sensitivity-Gated Accumulator Caching}
\label{sec:method}

SGAC solves the problem of Section~\ref{sec:problem} with three decisions. The deployed codes are the only copy of the convolution weights, so nothing is unfolded or re-made. Updates are measured in code steps: whole steps are written to the codes, and the fraction below one step is kept in an 8-bit accumulator. And only a small, changing subset of weight groups holds an accumulator at any time; a group that loses its accumulator writes back its remainder by a fair coin flip, and a group without one waits.

\subsection{Low-Bit Storage}
\label{sec:storage}

Every convolution is stored group-wise. A group $g$ holds $n_g=64$ weights with one scale $s_g$ and one offset $\beta_g$:
\begin{equation}
W_i=s_g\,q_i+\beta_g,\qquad q_i\in\{0,\dots,2^{b_l}-1\},
\label{eq:affine}
\end{equation}
where $l$ is the layer of weight $i$. The first convolution and the output head use $b_l=8$, and all other layers use $b_l=4$. These two layers hold less than 1\% of the weights, so the codes take 4.0 bits per weight on average, or 3.99 bits once the outliers below are stored separately (Table~\ref{tab:bits}); Section~\ref{sec:alloc} checks this choice. The 1\% of weights farthest from their group median are stored separately as outliers \cite{spqr}. In this format the keypoint network occupies 0.23\,MB (Fig.~\ref{fig:qual}(f)).

\subsection{Update in Code Units}

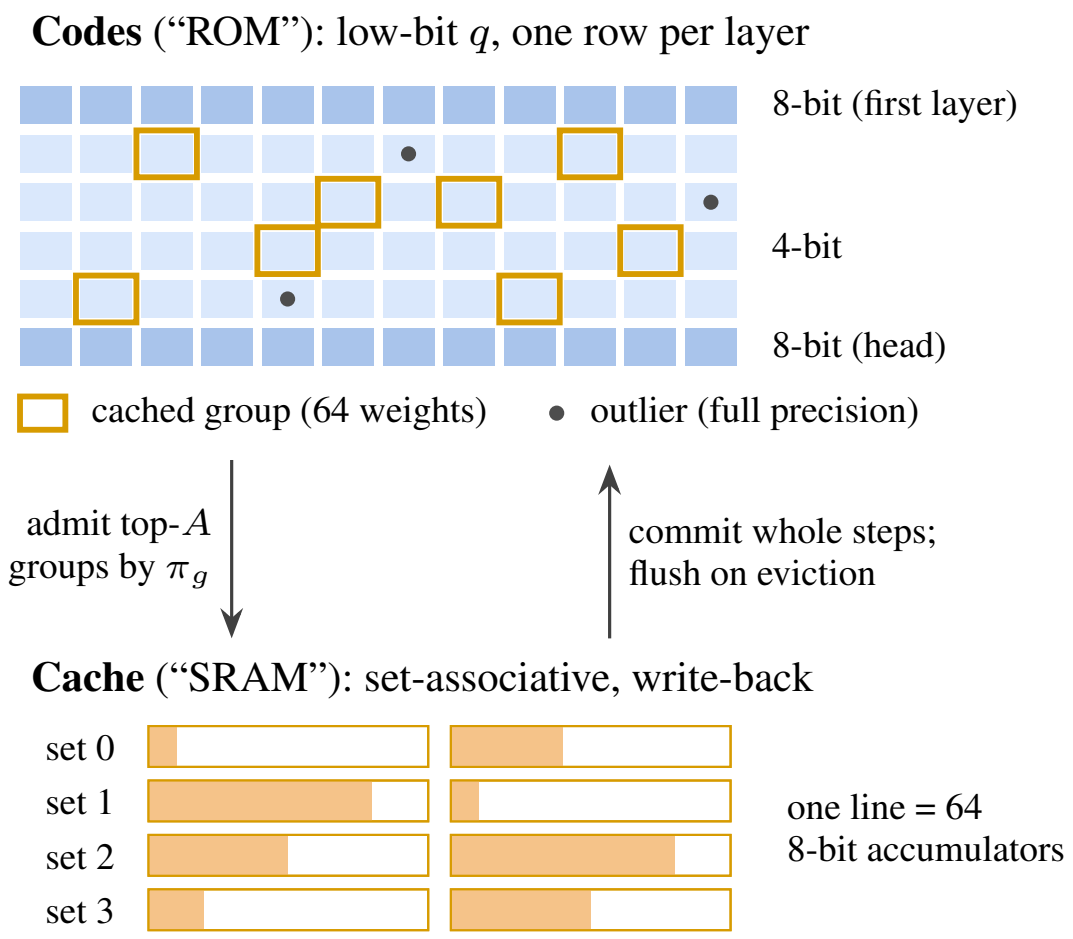
\begin{figure}[!tbp]
\centering
\adjustbox{max width=0.72\textwidth}{\definecolor{bF}{HTML}{DAE8FC}\definecolor{bD}{HTML}{6C8EBF}\definecolor{bM}{HTML}{A9C4EB}
\definecolor{oF}{HTML}{FFE6CC}\definecolor{oD}{HTML}{D79B00}\definecolor{oM}{HTML}{F5C389}
\begin{tikzpicture}[font=\footnotesize, >=Stealth, arr/.style={->, line width=0.6pt, draw=black!75}]
\node[anchor=south west] at (-0.05,0.34) {\textbf{Codes} (``ROM''): low-bit $q$, one row per layer};
\foreach \r in {0,...,5} { \foreach \c in {0,...,11} {
  \ifnum\r=0 \fill[bM] (\c*0.4, -\r*0.32) rectangle ++(0.34,0.25); \else \ifnum\r=5 \fill[bM] (\c*0.4, -\r*0.32) rectangle ++(0.34,0.25); \else \fill[bF] (\c*0.4, -\r*0.32) rectangle ++(0.34,0.25); \fi\fi } }
\node[anchor=west, font=\scriptsize] at (4.85,0.12) {8-bit (first layer)};
\node[anchor=west, font=\scriptsize] at (4.85,-0.80) {4-bit};
\node[anchor=west, font=\scriptsize] at (4.85,-1.48) {8-bit (head)};
\foreach \r/\c in {1/2, 2/7, 3/4, 4/1, 1/9, 3/10, 2/5, 4/8} { \draw[oD, line width=1.1pt] (\c*0.4-0.03, -\r*0.32-0.03) rectangle ++(0.40,0.31); }
\foreach \r/\c in {1/6, 4/4, 2/11} { \fill[black!70] (\c*0.4+0.17, -\r*0.32+0.125) circle (0.05); }
\draw[oD, line width=1.1pt] (0.0,-2.02) rectangle ++(0.30,0.22); \node[anchor=west, font=\scriptsize] at (0.35,-1.91) {cached group (64 weights)};
\fill[black!70] (3.55,-1.91) circle (0.05); \node[anchor=west, font=\scriptsize] at (3.65,-1.91) {outlier (full precision)};
\begin{scope}[shift={(0,-4.25)}]
\node[anchor=south west] at (-0.05,0.33) {\textbf{Cache} (``SRAM''): set-associative, write-back};
\foreach \s in {0,...,3} { \node[anchor=east, font=\scriptsize] at (0.75,-\s*0.36+0.13) {set \s};
  \foreach \w in {0,1} { \draw[oD, fill=white] (0.85+\w*2.0, -\s*0.36) rectangle ++(1.85,0.27);
    \pgfmathsetmacro{\fv}{1.85*mod(37*\s+53*\w+11,10)/10} \fill[oM] (0.85+\w*2.0, -\s*0.36) rectangle ++(\fv,0.27);
    \draw[oD] (0.85+\w*2.0, -\s*0.36) rectangle ++(1.85,0.27); } }
\node[anchor=west, font=\scriptsize, align=left] at (4.95,-0.40) {one line = 64\\8-bit accumulators};
\end{scope}
\draw[arr] (1.40,-2.22) -- node[left, font=\scriptsize, align=right]{admit top-$A$\\groups by $\pi_g$} (1.40,-3.40);
\draw[arr] (3.90,-3.40) -- node[right, font=\scriptsize, align=left]{commit whole steps;\\flush on eviction} (3.90,-2.25);
\end{tikzpicture}}
\caption{SGAC memory layout. The low-bit codes (``ROM'') are the only copy of the convolution weights; a set-associative cache (``SRAM'') holds 8-bit accumulators for a few groups, which write whole code steps back; uncached groups wait.}
\label{fig:sgac}
\end{figure}

SGAC solves (\ref{eq:p3}) by updating the codes of (\ref{eq:affine}) directly (Fig.~\ref{fig:sgac}, Algorithm~\ref{alg:sgac}). The backward pass runs through the low-bit weights and returns $g_i=\partial\mathcal{L}/\partial W_i$. The outliers and the biases, about 1\% of the parameters, are kept at full precision and trained with Adam; their state is counted in Table~\ref{tab:bits}.

\begin{figure}[!tbp]
\centering
\adjustbox{max width=0.72\textwidth}{
\begin{tikzpicture}[x=1pt,y=1pt,font=\footnotesize,
  cell/.style={draw=black!60,line width=0.3pt,minimum size=5.5pt,inner sep=0pt},
  on/.style={fill=blue!45},
  off/.style={fill=white},
  arr/.style={-{Latex[length=4pt]},line width=0.6pt},
  lbl/.style={font=\scriptsize,text=black!85},
  hd/.style={font=\scriptsize\bfseries,text=black!85}]

\newcommand{\grid}[3]{
  \foreach \m [count=\k from 0] in {#3}{
    \pgfmathtruncatemacro{\r}{int(\k/3)}
    \pgfmathtruncatemacro{\c}{mod(\k,3)}
    \ifnum\m=1
      \node[cell,on] at (#1+\c*6.5,#2-\r*6.5) {};
    \else
      \node[cell,off] at (#1+\c*6.5,#2-\r*6.5) {};
    \fi
  }}
\newcommand{\under}[2]{\node[lbl] at (#1+6.5,-6) {#2}; \draw[arr,black!50] (#1+6.5,-12) -- (#1+6.5,-22);}

\node[hd,anchor=west] at (0,30) {Training: six branches, each with batch normalization};
\grid{10}{18}{1,1,1,1,1,1,1,1,1}   \under{10}{$3{\times}3$}
\grid{56}{18}{0,0,0,1,1,1,0,0,0}   \under{56}{$1{\times}3$}
\grid{102}{18}{0,1,0,0,1,0,0,1,0}   \under{102}{$3{\times}1$}
\grid{148}{18}{0,0,0,0,1,0,0,0,0}  \under{148}{$1{\times}1$}
\grid{194}{18}{0,1,0,1,1,1,0,1,0}  \under{194}{``$+$'' stencil}
\grid{240}{18}{1,0,1,0,1,0,1,0,1}  \under{240}{``$\times$'' stencil}
\draw[line width=0.6pt,black!50] (16.5,-22) -- (246.5,-22);
\draw[arr] (131.5,-22) -- (131.5,-32);
\node[lbl,anchor=west] at (137,-27) {fold exactly: sum of the six kernels};
\grid{125}{-38}{1,1,1,1,1,1,1,1,1}
\node[lbl] at (131.5,-60) {deployed: one $3{\times}3$ kernel, stored as 4/8-bit codes};

\begin{scope}[yshift=-114pt]
\node[hd,anchor=west] at (0,30) {On the device: gradient $g$ of the folded kernel};
\grid{10}{18}{1,1,1,1,1,1,1,1,1}   \under{10}{$g$}
\grid{56}{18}{0,0,0,1,1,1,0,0,0}   \under{56}{$g\odot \Omega_r$}
\grid{102}{18}{0,1,0,0,1,0,0,1,0}   \under{102}{$g\odot \Omega_c$}
\grid{148}{18}{0,0,0,0,1,0,0,0,0}  \under{148}{$g\odot \Omega_o$}
\grid{194}{18}{0,1,0,1,1,1,0,1,0}  \under{194}{$\Phi_{+}\langle\Phi_{+},g\rangle$}
\grid{240}{18}{1,0,1,0,1,0,1,0,1}  \under{240}{$\Phi_{\times}\langle\Phi_{\times},g\rangle$}
\draw[line width=0.6pt,black!50] (16.5,-22) -- (246.5,-22);
\draw[arr] (131.5,-22) -- (131.5,-32);
\node[lbl,anchor=west] at (137,-27) {sum and rescale to $\|g\|$: (\ref{eq:compass})};
\grid{125}{-38}{1,1,1,1,1,1,1,1,1}
\node[lbl] at (131.5,-60) {$\hat g$: the direction the six branches would give, no branch stored};
\end{scope}
\end{tikzpicture}}
\caption{The compass gradient. A CompassConv block is trained with six branches and folded into one $3\times3$ kernel for deployment (top). On the device, the gradient $g$ of the folded kernel is projected onto the three masks and the two stencils of (\ref{eq:compass}) and summed (bottom), which gives the update direction the six branches would have produced, without storing a branch.}
\label{fig:compass}
\end{figure}
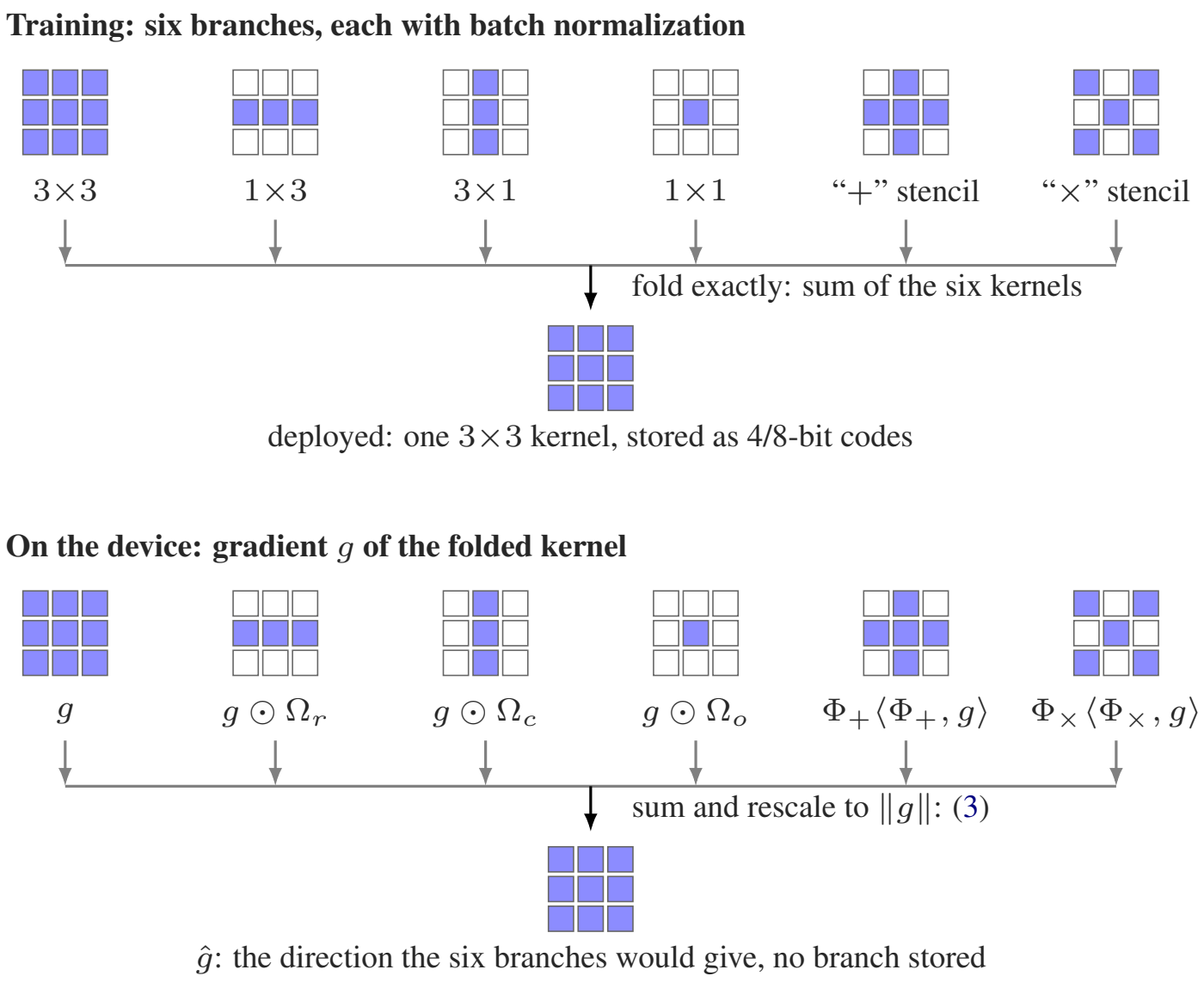

\textbf{Compass gradient.} Our networks are built from CompassConv blocks, which are trained as six branches and folded into one $3\times3$ kernel for deployment (Appendix~\ref{app:system}). For plain gradient descent, training several linear branches is equivalent to training the merged kernel with the gradient projected onto each branch and summed \cite{repopt,sgs}. For the CompassConv branches these projections are three masks and two rank-1 stencil projections (Fig.~\ref{fig:compass}), which give the direction
\begin{equation}
\hat g\propto g\odot(1+\Omega_r+\Omega_c+\Omega_o)+\Phi_{+}\langle \Phi_{+},g\rangle+\Phi_{\times}\langle \Phi_{\times},g\rangle,
\label{eq:compass}
\end{equation}
where $\Omega_r$, $\Omega_c$ and $\Omega_o$ are the row, column and center masks of the $3\times3$ kernel, $\Phi_{+}$ and $\Phi_{\times}$ are the axial and diagonal averaging stencils, and $\hat g$ is rescaled to the norm of $g$. The map is fixed and stores nothing, so SGAC trains the kernel the device actually runs instead of a set of branches kept beside it. In the 400-step adaptations of Section~\ref{sec:exp}, the compass and the plain gradient reach the same accuracy within 0.2 points (Table~\ref{tab:ablation}); the fold therefore costs nothing during adaptation.

\textbf{Update.} Each group keeps one exponential average of its squared gradient,
\begin{equation}
F_g\leftarrow\rho F_g+(1-\rho)\,\tfrac{1}{n_g}\textstyle\sum_{i\in g}\hat g_i^2,
\label{eq:fisher}
\end{equation}
and the update of weight $i$, in code steps, is
\begin{equation}
d_i=-\eta\,\frac{\hat g_i}{s_g\,(\sqrt{\hat F_g}+\epsilon)},
\label{eq:update}
\end{equation}
where $\hat F_g$ is the bias-corrected average and $\eta$ the learning rate. This is an RMSprop-style step with one second moment per group instead of one per weight. Most $|d_i|$ are far smaller than one code step, which is why a plain low-bit update would round them away.

\subsection{Write-Back Cache}

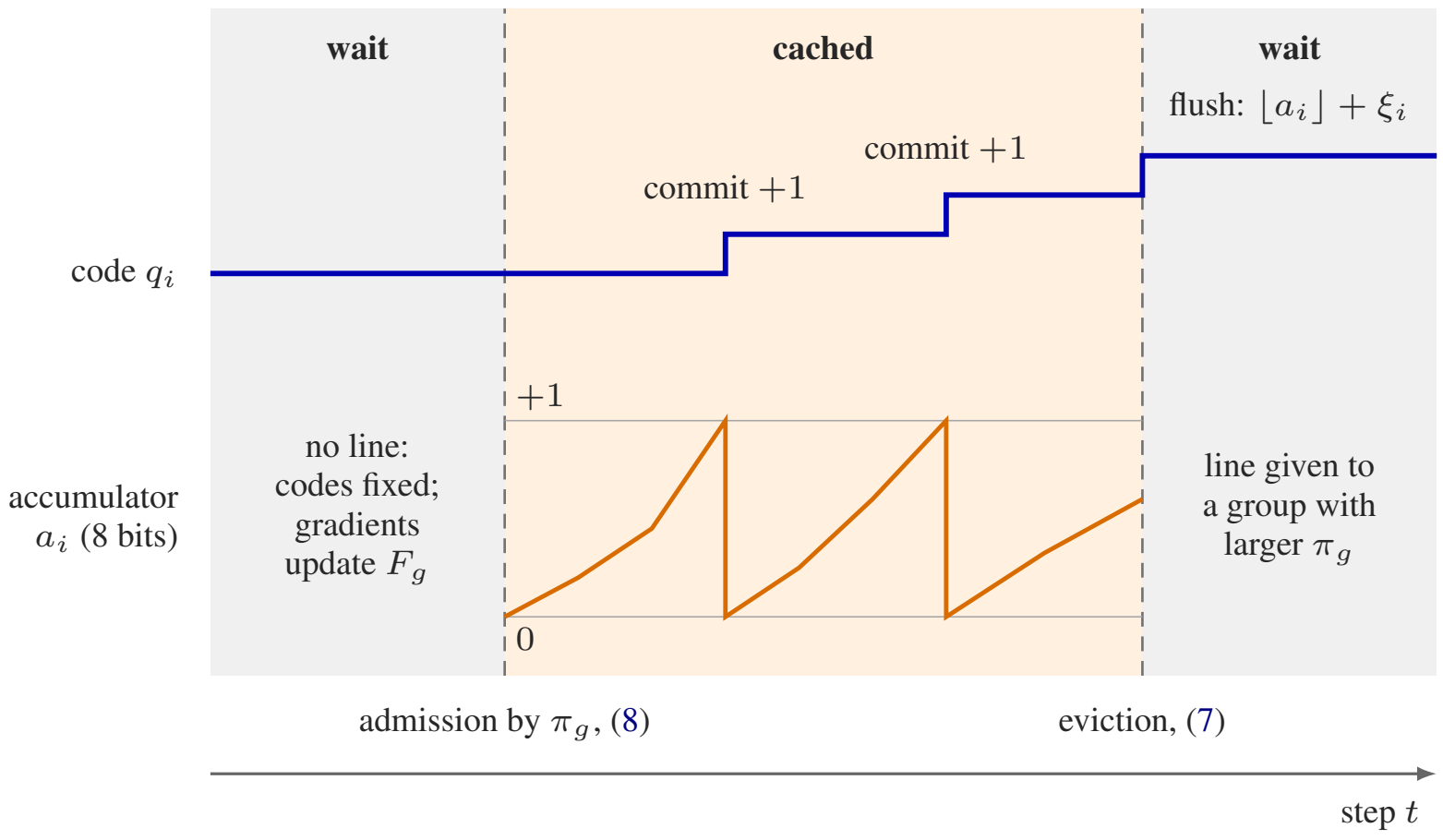
\begin{figure}[!tbp]
\centering
\adjustbox{max width=0.82\textwidth}{
\begin{tikzpicture}[x=1pt,y=1pt,font=\footnotesize,
  arr/.style={-{Latex[length=4pt]},line width=0.6pt},
  lbl/.style={font=\scriptsize,text=black!85},
  code/.style={line width=1.1pt,blue!70!black},
  acc/.style={line width=0.9pt,orange!85!black},
  ev/.style={densely dashed,black!55,line width=0.5pt}]

\fill[black!6]   (0,-72) rectangle (60,64);
\fill[orange!12] (60,-72) rectangle (190,64);
\fill[black!6]   (190,-72) rectangle (250,64);
\node[lbl] at (30,56) {\textbf{wait}};
\node[lbl] at (125,56) {\textbf{cached}};
\node[lbl] at (220,56) {\textbf{wait}};
\draw[ev] (60,-72) -- (60,64);
\draw[ev] (190,-72) -- (190,64);

\node[lbl,anchor=east] at (-3,10) {code $q_i$};
\draw[code] (0,10) -- (105,10) -- (105,18) -- (150,18) -- (150,26) -- (190,26) -- (190,34) -- (250,34);
\node[lbl,anchor=south] at (105,21) {commit $+1$};
\node[lbl,anchor=south] at (150,29) {commit $+1$};
\node[lbl,anchor=south] at (220,37) {flush: $\lfloor a_i\rfloor+\xi_i$};

\node[lbl,anchor=east,align=right] at (-3,-40) {accumulator\\$a_i$ (8 bits)};
\draw[black!35,line width=0.3pt] (60,-60) -- (190,-60);
\draw[black!35,line width=0.3pt] (60,-20) -- (190,-20);
\node[lbl,anchor=south west,inner sep=1pt] at (61,-19) {$+1$};
\node[lbl,anchor=north west,inner sep=1pt] at (61,-61) {$0$};
\draw[acc] (60,-60) -- (75,-52) -- (90,-42) -- (105,-20) -- (105,-60) -- (120,-50) -- (135,-36) -- (150,-20) -- (150,-60) -- (170,-47) -- (190,-36);

\node[lbl,align=center] at (30,-38) {no line:\\codes fixed;\\gradients\\update $F_g$};
\node[lbl,align=center] at (220,-38) {line given to\\a group with\\larger $\pi_g$};

\node[lbl,anchor=north] at (60,-75) {admission by $\pi_g$, (\ref{eq:priority})};
\node[lbl,anchor=north] at (190,-75) {eviction, (\ref{eq:evict})};
\draw[arr,black!60] (0,-92) -- (250,-92);
\node[lbl,anchor=north east] at (250,-94) {step $t$};
\end{tikzpicture}}
\caption{Wait-or-write-back for one weight group. While the group has no line, its codes do not move and its gradients only update $F_g$ (wait). Once admitted, its sub-step updates $d_i$ accumulate in 8 bits and the codes move by whole steps as the accumulator crosses $\pm1$ (commit). On eviction, the remainder (0.6 of a step in this example) is rounded stochastically into the codes (flush), so nothing is lost in expectation.}
\label{fig:lifecycle}
\end{figure}

The cache has $S$ sets of $A$ ways. Set-associativity limits the admission search to $A$ candidates per set; a direct-mapped and a fully associative cache give the same accuracy (Section~\ref{sec:ablation}), so the organization is a matter of cost. Fig.~\ref{fig:lifecycle} follows one group through the policy. A line holds the accumulators $a_i$ of one group and a tag, and each group maps to one set by a hash of its index. The sum $a_i+d_i$ is formed in a register, and only the remainder, $|a_i|<1$, is stored, in 8 bits with a resolution of $1/128$ of a code step (the remainder is stochastically rounded to this grid). For a cached group,
\begin{equation}
a_i\leftarrow a_i+d_i,\quad q_i\leftarrow q_i+\trunc(a_i),\quad a_i\leftarrow a_i-\trunc(a_i),
\label{eq:commit}
\end{equation}
so a code moves only by whole steps. A group without a line keeps its codes until it receives one; its gradients still update $F_g$, but its updates are not applied until it is admitted (\emph{deferred update}). When a line is evicted, its remainder is written back by stochastic rounding,
\begin{equation}
q_i\leftarrow q_i+\lfloor a_i\rfloor+\xi_i,\qquad \xi_i\sim\mathrm{Bernoulli}\big(a_i-\lfloor a_i\rfloor\big).
\label{eq:evict}
\end{equation}
If many codes of a group reach the end of their range, the group range is widened by one step and its codes are re-mapped.

\textbf{Admission.} Every $T$ steps, each set keeps the $A$ groups with the largest priority
\begin{equation}
\pi_g=n_g\,F_g\,s_g^2,
\label{eq:priority}
\end{equation}
the Fisher-weighted squared step that is used to choose bit widths \cite{fit,qbitopt}. Here it decides which groups receive accumulator precision. Section~\ref{sec:analysis} relates $\pi_g$ to the rounding noise a group would suffer without a line, and Section~\ref{sec:checks} measures how close the resulting cache comes to the best possible one.

\begin{algorithm}[!tbp]
\caption{One step of SGAC}
\label{alg:sgac}
\footnotesize
\begin{algorithmic}[1]
\Require codes $(q,s,\beta)$, cache with accumulators $a$, averages $F$, step $t$, batch $(x,y)$
\State Forward with $W=s\,q+\beta$ of (\ref{eq:affine}); backward to get $g=\partial\mathcal{L}/\partial W$
\State $\hat g\gets$ compass gradient of $g$ by (\ref{eq:compass}); update $F_g$ by (\ref{eq:fisher})
\For{each cached group $g$ and each $i\in g$}
 \State $d_i\gets$ (\ref{eq:update}); commit whole steps by (\ref{eq:commit})
\EndFor
\State Uncached groups: codes unchanged (wait)
\State Update outliers and biases with Adam
\If{$t \bmod T=0$}
 \State $\pi_g\gets n_gF_gs_g^2$ by (\ref{eq:priority}) for every group
 \State In each set, keep the top-$A$ groups; flush evicted lines by (\ref{eq:evict}); set $a\gets0$ for new lines
\EndIf
\State \Return updated codes and cache (the codes are the deployable network)
\end{algorithmic}
\end{algorithm}

\subsection{Training Memory}

With $\bar b$-bit codes, 16-bit scales and offsets and a 32-bit $F_g$ per group of 64 weights, a cache of $B$ bits per weight, and a share $p_o$ of outliers stored with a 16-bit value, a 16-bit index and two Adam moments, the persistent training state per weight is
\begin{equation}
M\approx\bar b+1+B+96\,p_o+m_b\ \text{bits},
\label{eq:memory}
\end{equation}
where $m_b$ counts the biases and their Adam moments. Table~\ref{tab:bits} lists every component for the main methods. For SGAC, (\ref{eq:memory}) gives 8.1 bits per weight, of which 2 are the cache; with 8-bit lines of 64 accumulators and a 32-bit tag, $B=2$ covers 23\% of the groups. A 32-bit master copy with Adam needs 96 bits.

\begin{table}[!tbp]
\caption{Persistent training state per weight (bits).}
\label{tab:bits}
\centering
\small
\setlength{\tabcolsep}{2.4pt}
\begin{adjustbox}{max width=\textwidth}
\begin{threeparttable}
\begin{tabular}{@{}lccccc@{}}
\toprule
Component & FP + Adam & QAT & QLoRA-4 & Direct SR & \textbf{SGAC} \\
\midrule
weights (32-bit) or master copy & 32 & 32 & -- & -- & -- \\
Adam moments of the weights & 64 & 64 & -- & -- & -- \\
low-bit codes ($\bar b$) & -- & -- & 3.99 & 3.99 & 3.99 \\
scales and offsets (16-bit per 64) & -- & -- & 0.52 & 0.52 & 0.52 \\
adapters + Adam (rank 4) & -- & -- & 6.48 & -- & -- \\
gradient averages $F_g$ (32-bit per 64) & -- & -- & -- & 0.52 & 0.52 \\
accumulator cache ($B$) & -- & -- & -- & -- & 2.00 \\
outliers: value + index & -- & -- & 0.32 & 0.32 & 0.32 \\
outliers: Adam moments & -- & -- & -- & 0.64 & 0.64 \\
biases + Adam & -- & -- & 0.16 & 0.16 & 0.16 \\
\midrule
\textbf{total} & 96 & 96 & 11.47 & 6.14 & \textbf{8.14} \\
\bottomrule
\end{tabular}
\begin{tablenotes}[flushleft]\footnotesize
\item[] Outliers (1\% of the weights) and biases are trained with Adam by direct SR and SGAC; QLoRA keeps the codes, scales, offsets and outliers frozen and trains its adapters and the biases, so it carries no Adam state for the outliers. Ranks 1 and 8 of QLoRA give 6.61 and 17.95 bits.
\end{tablenotes}
\end{threeparttable}
\end{adjustbox}
\end{table}

\section{Analysis: When Waiting Beats Rounding}
\label{sec:analysis}

This section explains the two halves of the policy. Proposition~\ref{prop:unbiased} shows that the cache changes the noise of learning but never its direction. Propositions~\ref{prop:first} and~\ref{prop:wait} show what rounding costs and when a group without a line is better off waiting, and Corollary~\ref{cor:admission} shows which groups should hold a line.

\textbf{Assumptions.} Proposition~\ref{prop:unbiased} is exact. The other statements describe a single step and rest on four assumptions. (A1)~The loss is replaced by its second-order Taylor model around the current weights. (A2)~Rounding errors are independent across weights and of the batch; stochastic rounding satisfies this by construction. (A3)~Off-diagonal curvature terms are dropped; they enter at order $\eta^2$, and the device has no access to them in any case. (A4)~Within a step, the group average $F_g$ is treated as a constant. Section~\ref{sec:checks} tests the predictions numerically on a quadratic and on PalmNet-K. Throughout, $g$ denotes the gradient that drives the update; in a CompassConv layer this is the compass gradient $\hat g$ of (\ref{eq:compass}).

Throughout, an update of $d_i$ code steps with $|d_i|<1$ is applied in one of three ways: \emph{exactly} (the group is cached, and the update is accumulated and written back in whole steps), by \emph{stochastic rounding} (the code moves by $\xi_i\in\{\lfloor d_i\rfloor,\lfloor d_i\rfloor+1\}$ with $\mathbb{E}[\xi_i]=d_i$), or not at all (\emph{wait}).

\begin{proposition}[Unbiased write-back]\label{prop:unbiased}
Let $u_t$ be the update (\ref{eq:update}) of one weight at step $t$ if its group holds a line, and $u_t=0$ otherwise, and let $a_t$ be its remainder. For any admission and eviction schedule and any horizon $t_1$, $q_{t_1}+a_{t_1}-q_0=\sum_{t<t_1}u_t+Z_{t_1}$, where $Z_{t_1}$ is a zero-mean martingale, up to clipping and re-ranging of the code range.
\end{proposition}
\begin{proof}
A commit (\ref{eq:commit}) moves an amount from $a$ to $q$ and leaves $q+a$ unchanged. The stochastic rounding of the remainder to its 8-bit grid, and an eviction (\ref{eq:evict}), which sets $a$ to zero and moves $q$ by $\lfloor a\rfloor+\xi$ with $\mathbb{E}[\xi]=a-\lfloor a\rfloor$, each change $q+a$ by an amount with zero mean given the past. Summing over steps gives the result.
\end{proof}

\begin{proposition}[The rounding cost is first order in the step]\label{prop:first}
Let $\ell$ be a loss with gradient $G$ and Hessian $H$ at the current weights. Under (A1)--(A3), if the codes of a group move by stochastic rounding of $d$, with $|d_i|<1$, instead of exactly by $d$, the expected loss exceeds that of the exact move by
\begin{equation}
\mathbb{E}[\Delta\ell_{\mathrm{SR}}]-\Delta\ell_{\mathrm{exact}}=\tfrac12\,s_g^2\textstyle\sum_{i\in g}H_{ii}\,|d_i|\,(1-|d_i|),
\label{eq:first}
\end{equation}
a quantity that is of first order in $|d_i|$ for small steps.
\end{proposition}
\begin{proof}
Write $\xi_i=d_i+e_i$. The rounding errors $e_i$ are independent with zero mean, so the first-order term in $e$ and the off-diagonal second-order terms vanish in expectation, and $\mathbb{E}[\Delta\ell_{\mathrm{SR}}]=\Delta\ell_{\mathrm{exact}}+\tfrac12 s_g^2\sum_i H_{ii}\,\mathrm{Var}(\xi_i)$. For $|d_i|<1$, $\xi_i$ takes two adjacent integer values and $\mathrm{Var}(\xi_i)=|d_i|(1-|d_i|)$.
\end{proof}

The exact move changes the loss by $s_g\sum_i G_i d_i$, which is also linear in $d$, plus a curvature term of order $d^2$. The rounding cost (\ref{eq:first}) is therefore of the same order as the gain of the move. Halving the learning rate halves both, so for small steps the noise never becomes negligible relative to the progress. Since $|d|(1-|d|)\le1/4$, the cost per step is at most $\tfrac18 s_g^2\sum_i H_{ii}$.

\textbf{Which objective.} The deployed network has to serve the new condition without giving up the old one: a user moves between rooms and cameras, and a calibration that ruins accuracy elsewhere is not acceptable. We therefore analyze the weighted objective
\begin{equation}
J_\lambda=\lambda\,\ell'+(1-\lambda)\,\ell,\qquad \lambda\in[0,1],
\label{eq:jlambda}
\end{equation}
where $\ell'$ and $\ell$ are the losses on the new and the original distribution and $\lambda$ is the share of use the device expects in the new condition. Our evaluation reports both test sets and their mean, which corresponds to $\lambda=\tfrac12$ in (\ref{eq:jlambda}); Section~\ref{sec:checks} repeats the measurement for $\lambda$ from 0 to 1, and Section~\ref{sec:main} shows how the ranking of the methods changes with the weight.

\begin{proposition}[Wait or round]\label{prop:wait}
Let the update be (\ref{eq:update}), driven by batch gradients $g$ of $\ell'$ with $\mathbb{E}[g_i]=G'_i$, and let $G',G$ and $H',H$ be the gradients and Hessians of $\ell'$ and $\ell$. Under (A1)--(A4) and up to terms of order $\eta^2$, rounding a group instead of letting it wait changes $\mathbb{E}[J_\lambda]$ by
\begin{equation}
\frac{\eta}{\sqrt{F_g}}\Big[-\textstyle\sum_i G^\lambda_i G'_i+\tfrac{s_g}{2}\sum_i H^\lambda_{ii}\,\mathbb{E}|g_i|\Big],
\label{eq:wait}
\end{equation}
with $G^\lambda=\lambda G'+(1-\lambda)G$ and $H^\lambda=\lambda H'+(1-\lambda)H$. Waiting is better whenever
\begin{equation}
r_g(\lambda)=\frac{2\sum_i G^\lambda_i\,G'_i}{s_g\sum_i H^\lambda_{ii}\,\mathbb{E}|g_i|}<1.
\label{eq:ratio}
\end{equation}
The condition does not depend on $\eta$. If the network has converged on $\mathcal{D}$, so that $G\approx0$, rounding can only increase the original loss, by $\tfrac{\eta s_g}{2\sqrt{F_g}}\sum_i H_{ii}\mathbb{E}|g_i|>0$.
\end{proposition}
\begin{proof}
Waiting leaves $J_\lambda$ unchanged. Rounding moves code $i$ by $\xi_i$ with $\mathbb{E}[\xi_i\mid g]=d_i$ and, for $|d_i|<1$, $\mathbb{E}[\xi_i^2\mid g]=|d_i|$. Expanding $J_\lambda$ to second order in the weight change $s_g\xi$ and taking expectations over the rounding and over the batch gives
$\mathbb{E}[\Delta J_\lambda]=s_g\sum_i G^\lambda_i\,\mathbb{E}[d_i]+\tfrac12 s_g^2\sum_i H^\lambda_{ii}\,\mathbb{E}|d_i|+O(\eta^2)$;
the off-diagonal terms are of order $\eta^2$ because $\mathbb{E}[\xi_i\xi_j]=d_id_j$ for $i\ne j$. Substituting $\mathbb{E}[d_i]=-\eta G'_i/(s_g\sqrt{F_g})$ and $\mathbb{E}|d_i|=\eta\,\mathbb{E}|g_i|/(s_g\sqrt{F_g})$ gives (\ref{eq:wait}).
\end{proof}

Proposition~\ref{prop:wait} separates two effects that a single loss would hide. On the new data, rounding a group with a strong and consistent gradient can beat waiting. On the original data, where the update brings little or nothing, rounding costs loss at first order; this is forgetting caused by the arithmetic rather than by the data. A cached group gets the first-order gain without the rounding noise. For $\lambda=1$ the ratio (\ref{eq:ratio}) involves the new loss alone, and for $\lambda=\tfrac12$ it is the ratio we measure on PalmNet-K in Section~\ref{sec:checks}.

\textbf{Which groups should get a line.} Give each group one of three treatments, cache, round or wait, with at most $A$ cached groups per set. To first order in $\eta$, the expected change of $J_\lambda$ separates over groups, the cached groups cost nothing beyond the exact update (up to the $1/128$ resolution of the accumulator), and the treatment of one group does not change the cost of another. Hence:

\begin{corollary}[Assignment]\label{cor:admission}
Under (A1)--(A4) and up to $O(\eta^2)$, an assignment minimizes $\mathbb{E}[J_\lambda]$ if and only if (i)~every uncached group waits when $r_g(\lambda)<1$ and is rounded when $r_g(\lambda)>1$, and (ii)~each set caches the (at most) $A$ groups with the largest positive line value $V_g=\frac{\eta}{\sqrt{F_g}}\min\big(\sum_i G^\lambda_iG'_i,\ \tfrac{s_g}{2}\sum_i H^\lambda_{ii}\mathbb{E}|g_i|\big)$, the gain of a line over the better uncached treatment. If the uncached groups are rounded, as in direct low-bit training, the line value is the rounding noise (\ref{eq:first}), which is bounded by $\tfrac18\pi_g$ when $H_{ii}\approx F_g$; per-set admission by $\pi_g$ then minimizes this upper bound.
\end{corollary}
\begin{proof}
The objective is a sum of per-group terms, each depending only on that group's treatment. For a fixed cached set, (i) minimizes each uncached term, and the remaining choice is a partition matroid (at most $A$ per set), so a greedy per-set choice of the $A$ largest savings $V_g$ is optimal.
\end{proof}

The device does not know $G$, $G'$ or $H$; it tracks only $F_g$. SGAC therefore admits by $\pi_g$ and lets every uncached group wait. Corollary~\ref{cor:admission} describes what this approximates: the rule is exact for the rounding-noise bound, and Section~\ref{sec:checks} measures how much of the ideal line value $\sum_g V_g$ the $\pi_g$-ranked cache actually holds.

\section{Experiments}
\label{sec:exp}

\begin{figure}[!tbp]
\centering
\includegraphics[width=0.96\textwidth]{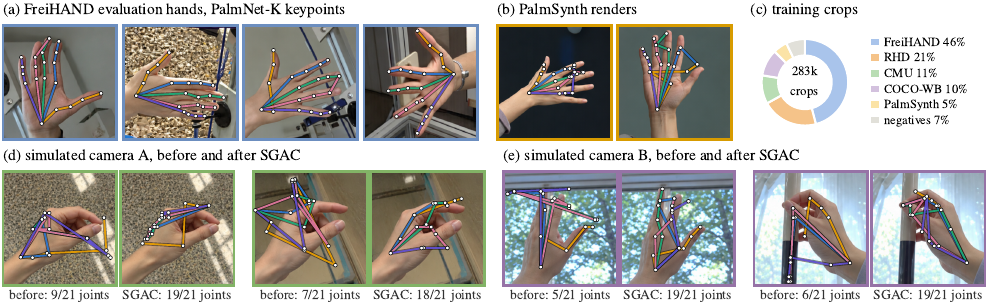}\\[4pt]
\resizebox{0.70\textwidth}{!}{\input{figs/fig_palmnetk.tex}}\\[-2pt]
{\footnotesize (f) PalmNet-K: a compact U-Net stored at 4/8 bits}
\caption{Qualitative results and the keypoint network. (a)--(b) keypoints of the final PalmNet-K; (c) training crops; (d)--(e) the same evaluation hands through two simulated cameras before adaptation and after SGAC (PalmNet-K$_0$), with the number of joints within PCK@0.2 (images enhanced for display; keypoints are network outputs); (f) architecture.}
\label{fig:qual}
\end{figure}

\subsection{Setup}

\textbf{Network and data.} All 400-step adaptation experiments start from the same checkpoint of PalmNet-K, a 383k-parameter U-Net that predicts 21 keypoint heatmaps from a $96\times96$ crop (Fig.~\ref{fig:qual}(f)), trained from scratch on FreiHAND \cite{freihand}, RHD \cite{zimmermann2017}, CMU hands \cite{simon2017}, COCO-WholeBody hands \cite{cocowholebody} and our PalmSynth renders (Table~\ref{tab:data}). We test adaptation under three kinds of shift. \emph{Camera~A} is a dark, warm, noisy webcam (gamma 1.6--2.4, 55--75\% brightness, a color cast, blur and sensor noise); \emph{camera~B} is a bright, cool, low-resolution camera with horizontal motion blur and coarse tone steps. For each camera the network adapts on 12,000 FreiHAND training crops seen through it and is tested on 1,000 FreiHAND evaluation hands, both through the camera (the new domain) and as original images, which measures forgetting. \emph{FreiHAND-C} applies the common corruptions of ImageNet-C \cite{hendrycks2019}, the protocol behind the COCO-C pose benchmark \cite{wang2021cococ}, to the same crops; Section~\ref{sec:fhc} gives the details. Real data complete the picture: in-the-wild COCO-WholeBody photographs, the domain on which both hand networks are weakest although it is part of their training data (Sections~\ref{sec:real} and~\ref{sec:second}), a webcam and a user that no network has seen (Section~\ref{sec:real}), and a new microphone for a sound network (Section~\ref{sec:second}).

\begin{table}[!tbp]
\caption{Datasets, the networks they train or test, and their sizes.}
\label{tab:data}
\centering
\small
\setlength{\tabcolsep}{5pt}
\begin{adjustbox}{max width=\textwidth}
\begin{tabular}{@{}llr@{}}
\toprule
Dataset & Used for & Train / test \\
\midrule
FreiHAND \cite{freihand} & PalmNet-K training; all adaptation and test sets of Sec.~\ref{sec:main}--\ref{sec:checks}; & 130,240 / 3,960 \\
 & hands cut out for the finder's composites & \\
RHD \cite{zimmermann2017} & PalmNet-K training; finder training and test & 59,592 / 3,753 \\
COCO-WholeBody hands \cite{cocowholebody} & PalmNet-K training; target domain in Sec.~\ref{sec:real} & 28,575 / 1,300 \\
COCO-WholeBody frames \cite{cocowholebody} & finder training; target domain in Sec.~\ref{sec:second} & train / val.\ frames \\
CMU hands \cite{simon2017} & PalmNet-K training & 30,249 / 846 \\
PalmSynth (ours) & PalmNet-K training and hand presence; finder frames & 15,000 / 1,500 \\
Places365 \cite{places} & room backgrounds for the finder's composites & room photos \\
HaGRID \cite{hagrid} & finder training and test users; finger states for the wide PalmNet-K & train subset / test users \\
HInt \cite{hamer} & wide PalmNet-K training and test (App.~\ref{app:results}) & official splits \\
Generated hand images (ours) & rare hand shapes (App.~\ref{app:results}) & 693 images \\
User's webcam recording (ours) & real camera and unseen user (Sec.~\ref{sec:real}) & 281 / 140 crops \\
Two-hand streams (ours) & gestures & 160 + 24 cal.\ / 30 + 10 min \\
ESC-50 \cite{esc50} & SoundTag & 2,000 clips, 5 folds \\
\bottomrule
\end{tabular}
\end{adjustbox}
\end{table}

\textbf{Checkpoints.} Each network appears in an adaptation checkpoint and a final version, and every accuracy in the paper is reported with its precision. \emph{PalmNet-K$_0$}, the adaptation checkpoint, is the starting point of Sections~\ref{sec:main}--\ref{sec:rivals}, \ref{sec:ablation} and~\ref{sec:checks}, of the real-camera study and of the pose-prior evaluation; on the 1,000 FreiHAND evaluation hands it reaches a PCK@0.2 of 0.679 in full precision and 0.680 as deployed 4/8-bit codes. The final \emph{PalmNet-K}, the same architecture from a later training round (Table~\ref{tab:keypoints}), reaches 0.728 in full precision, 0.625 as 4/8-bit codes quantized after training, which are the starting point of the longer adaptation in Section~\ref{sec:real}, and 0.736 after 12 minutes of quantization-aware training at uniform 4 bits (Section~\ref{sec:alloc}). The hand finder has an adaptation checkpoint \emph{PalmNet-D$_0$} (Sections~\ref{sec:second} and~\ref{sec:alloc}) and a final network \emph{PalmNet-D} (Table~\ref{tab:mediapipe}); a wide 1.1M-parameter PalmNet-K appears only in Appendix~\ref{app:results}.

\textbf{Metrics.} Keypoint accuracy is the percentage of correct keypoints,
\begin{equation}
\mathrm{PCK}@\theta=\frac{1}{|\mathcal{V}|}\sum_{(n,j)\in\mathcal{V}}\mathbf{1}\big[\|\hat p_{nj}-p_{nj}\|\le\theta L\big],
\label{eq:pck}
\end{equation}
where $\mathcal{V}$ is the set of visible keypoints of the test hands and $L$ is the crop size; a \emph{point} is 0.01 of PCK. For adaptation we report PCK@0.2 of (\ref{eq:pck}) on the new domain, on the original images, and their \emph{average}, which weighs learning and forgetting equally. Training memory is the persistent state $M$ of (\ref{eq:memory}); 1\,KB is 1,000 bytes throughout. Activation memory is the same for all methods and is reduced by orthogonal techniques \cite{tinytl,ondevice256}.

\textbf{What is measured.} Problem (\ref{eq:p3}) asks for a better low-bit network, so every method is scored on the network the device stores and runs after adaptation, in the same 4/8-bit layout as before. SGAC, QAT, direct low-bit training and the PEQA- and ECO-style baselines end in such codes by construction. A full-precision model is quantized after training with the rule that produced the starting codes. QLoRA adapters are merged into the weights and re-quantized, on the unchanged grid or on a re-fitted grid, whichever is better on validation data. Most low-bit training studies report full-precision weights or kept adapters; we also report QLoRA with its adapters kept, which is not a 4/8-bit network, to show what this choice hides.

\textbf{Baselines and protocol.} We compare SGAC with full-precision fine-tuning with Adam, QAT with a 32-bit master copy and a straight-through estimator \cite{jacob2018,lsq}, QLoRA with rank-1, 4 or 8 adapters on every convolution of the same 4/8-bit network, whose codes, scales, offsets and outliers stay frozen \cite{qlora}, a PEQA-style baseline that learns only the per-group scales and offsets \cite{peqa}, an ECO-style baseline that keeps a 16-bit momentum per weight and feeds the rounding error back into it \cite{eco}, and direct low-bit training by stochastic rounding \cite{gupta2015,smgd,dqt}. Section~\ref{sec:rivals} adds QA-LoRA, LoQT and SpIEL. All methods use 400 steps with batch size 32 and the same bit width per layer. The step size of every method and, for QLoRA, the merge mode are chosen on a validation split that is never used for adaptation or testing: 1,000 crops of a separate FreiHAND training file, through the camera and as original images. Selection uses the average validation accuracy over both cameras, so each method keeps a single setting. A three-seed screen first narrowed the candidates ($\{1/3,1,3\}\times1.5\times10^{-4}$ for full precision, $\{2,4,8,16\}\times$ for LoRA, $\{1,2,4\}\times$ for SGAC, and three values each for the scale step of PEQA and the step size of ECO), and the remaining candidates were run with all seeds. SGAC uses a four-way cache with $B=2$, admission every $T=25$ steps, $\rho=0.99$ and $\epsilon=10^{-8}$. FreiHAND-C reuses these settings without re-tuning. Step times are measured on one CPU thread (Intel Xeon, 2.1\,GHz).

\textbf{Statistics.} Table~\ref{tab:main} uses ten seeds per camera for SGAC, full precision and QLoRA with kept adapters, and five seeds for the other settings. A seed fixes the data order, so methods are paired by seed, and each paired difference is computed over the seeds the two methods share. We report paired 95\% confidence intervals (CIs) and paired $t$-tests. Each table is compared within itself: SGAC's camera-A accuracy on the original images is 0.658 in Table~\ref{tab:main} (validated step size, ten seeds), 0.640 in Table~\ref{tab:rivals} (the same setting in separate CPU runs with their own data order) and 0.667 in Table~\ref{tab:ablation} (the smaller common step size, which forgets less). We call two methods \emph{equivalent} when a two one-sided test (TOST) \cite{tost} places their paired difference within $\pm0.5$ points at the 5\% level. The margin is about 1.5 times the seed-to-seed standard deviation of SGAC's own average (0.28--0.34 points), so a difference inside it is smaller than what a new seed changes. The paired standard deviations of the differences in Table~\ref{tab:main} lie between 0.2 and 0.7 points; at 0.5 points, five seeds detect a 1-point difference with 80\% power, and a 0.6-point difference needs eight. Where fewer seeds are available, we report the interval and draw no conclusion it does not support.

\subsection{Learning in Low-Bit Memory}
\label{sec:main}

\begin{table}[!tbp]
\caption{Deployed 4/8-bit PalmNet-K$_0$ after adaptation to two simulated cameras (PCK@0.2, mean $\pm$ std over seeds).}
\label{tab:main}
\centering
\small
\setlength{\tabcolsep}{2.0pt}
\begin{adjustbox}{max width=\textwidth}
\begin{threeparttable}
\begin{tabular}{@{}lccccc@{}}
\toprule
Method & \begin{tabular}[c]{@{}c@{}}State\\(bits/w)\end{tabular} & New camera & Original & Average & \begin{tabular}[c]{@{}c@{}}SGAC$-$\\method (pts)\end{tabular} \\
\midrule
\multicolumn{6}{@{}l}{\textit{Camera A}} \\
Before adaptation, full precision & -- & 0.558 & 0.679 & 0.618 & -- \\
Before adaptation, 4/8-bit codes & -- & 0.545 & 0.680 & 0.612 & -- \\
Full precision + Adam, then 4/8-bit & 96 & \pmstd{0.587}{0.007} & \pmstd{0.669}{0.004} & \pmstd{\textbf{0.628}}{0.004} & $-$0.1 \\
QAT, 32-bit master copy & 96 & \pmstd{\textbf{0.596}}{0.007} & \pmstd{0.630}{0.007} & \pmstd{0.613}{0.005} & +1.4$^*$ \\
QLoRA rank 1, merged & 6.61 & \pmstd{0.583}{0.005} & \pmstd{0.656}{0.009} & \pmstd{0.619}{0.006} & +0.8$^*$ \\
QLoRA rank 4, merged & 11.47 & \pmstd{0.589}{0.005} & \pmstd{0.658}{0.006} & \pmstd{0.623}{0.004} & +0.4 \\
QLoRA rank 8, merged & 17.95 & \pmstd{0.584}{0.005} & \pmstd{0.656}{0.008} & \pmstd{0.620}{0.004} & +0.7$^*$ \\
PEQA-style (grid only) & 8.72 & \pmstd{0.582}{0.005} & \pmstd{0.654}{0.004} & \pmstd{0.618}{0.003} & +0.9$^*$ \\
ECO-style (momentum feedback) & 21.98 & \pmstd{0.570}{0.002} & \pmstd{\textbf{0.679}}{0.002} & \pmstd{0.624}{0.002} & +0.3 \\
Direct low-bit (SR) & 6.14 & \pmstd{0.564}{0.015} & \pmstd{0.632}{0.007} & \pmstd{0.598}{0.009} & +2.9$^*$ \\
\textbf{SGAC (ours)} & \textbf{8.14} & \pmstd{0.595}{0.003} & \pmstd{0.658}{0.004} & \pmstd{0.627}{0.003} & -- \\
\textit{QLoRA rank 4, adapters kept$^\dagger$} & 11.47 & \pmstd{0.588}{0.003} & \pmstd{0.671}{0.003} & \pmstd{0.629}{0.002} & $-$0.3$^*$ \\
\midrule
\multicolumn{6}{@{}l}{\textit{Camera B}} \\
Before adaptation, full precision & -- & 0.445 & 0.679 & 0.562 & -- \\
Before adaptation, 4/8-bit codes & -- & 0.430 & 0.680 & 0.555 & -- \\
Full precision + Adam, then 4/8-bit & 96 & \pmstd{0.593}{0.005} & \pmstd{0.648}{0.006} & \pmstd{0.620}{0.004} & +1.1$^*$ \\
QAT, 32-bit master copy & 96 & \pmstd{0.613}{0.012} & \pmstd{0.624}{0.007} & \pmstd{0.619}{0.009} & +1.3$^*$ \\
QLoRA rank 1, merged & 6.61 & \pmstd{0.597}{0.007} & \pmstd{0.661}{0.007} & \pmstd{0.629}{0.003} & +0.2 \\
QLoRA rank 4, merged & 11.47 & \pmstd{0.560}{0.010} & \pmstd{0.677}{0.004} & \pmstd{0.618}{0.004} & +1.3$^*$ \\
QLoRA rank 8, merged & 17.95 & \pmstd{0.530}{0.008} & \pmstd{\textbf{0.683}}{0.001} & \pmstd{0.607}{0.004} & +2.4$^*$ \\
PEQA-style (grid only) & 8.72 & \pmstd{\textbf{0.619}}{0.002} & \pmstd{0.652}{0.005} & \pmstd{\textbf{0.636}}{0.003} & $-$0.5$^*$ \\
ECO-style (momentum feedback) & 21.98 & \pmstd{0.556}{0.001} & \pmstd{0.672}{0.002} & \pmstd{0.614}{0.001} & +1.7$^*$ \\
Direct low-bit (SR) & 6.14 & \pmstd{0.580}{0.005} & \pmstd{0.614}{0.013} & \pmstd{0.597}{0.005} & +3.4$^*$ \\
\textbf{SGAC (ours)} & \textbf{8.14} & \pmstd{0.618}{0.004} & \pmstd{0.646}{0.004} & \pmstd{0.632}{0.003} & -- \\
\textit{QLoRA rank 4, adapters kept$^\dagger$} & 11.47 & \pmstd{0.618}{0.002} & \pmstd{0.651}{0.003} & \pmstd{0.634}{0.002} & $-$0.2$^*$ \\
\midrule
\multicolumn{6}{@{}l}{\textit{Both cameras (per-seed mean of A and B); last column: SGAC$-$method with paired 95\% CI}} \\
Full precision + Adam, then 4/8-bit & 96 & \pmstd{0.590}{0.004} & \pmstd{0.658}{0.004} & \pmstd{0.624}{0.003} & $+$0.5 [$+$0.3, $+$0.7] \\
QAT, 32-bit master copy & 96 & \pmstd{0.604}{0.005} & \pmstd{0.627}{0.006} & \pmstd{0.616}{0.005} & $+$1.3 [$+$0.7, $+$2.0] \\
QLoRA rank 1, merged & 6.61 & \pmstd{0.590}{0.004} & \pmstd{0.658}{0.007} & \pmstd{0.624}{0.004} & $+$0.5 [$+$0.3, $+$0.7] \\
QLoRA rank 4, merged & 11.47 & \pmstd{0.574}{0.005} & \pmstd{0.667}{0.003} & \pmstd{0.621}{0.004} & $+$0.8 [$+$0.3, $+$1.4] \\
QLoRA rank 8, merged & 17.95 & \pmstd{0.557}{0.005} & \pmstd{0.669}{0.004} & \pmstd{0.613}{0.003} & $+$1.6 [$+$1.2, $+$1.9] \\
PEQA-style (grid only) & 8.72 & \pmstd{0.601}{0.003} & \pmstd{0.653}{0.003} & \pmstd{0.627}{0.003} & $+$0.2 [$+$0.1, $+$0.4] \\
ECO-style (momentum feedback) & 21.98 & \pmstd{0.563}{0.001} & \pmstd{0.676}{0.001} & \pmstd{0.619}{0.001} & $+$1.0 [$+$0.8, $+$1.2] \\
Direct low-bit (SR) & 6.14 & \pmstd{0.572}{0.009} & \pmstd{0.623}{0.007} & \pmstd{0.598}{0.003} & $+$3.1 [$+$2.6, $+$3.7] \\
\textbf{SGAC (ours)} & \textbf{8.14} & \pmstd{0.606}{0.003} & \pmstd{0.652}{0.003} & \pmstd{0.629}{0.002} & -- \\
\textit{QLoRA rank 4, adapters kept$^\dagger$} & 11.47 & \pmstd{0.603}{0.001} & \pmstd{0.661}{0.002} & \pmstd{0.632}{0.001} & $-$0.3 [$-$0.4, $-$0.1] \\
\bottomrule
\end{tabular}
\begin{tablenotes}[flushleft]\footnotesize
\item[] Average: mean of the new camera and the original images. Every method starts from the deployed 4/8-bit codes; the full-precision network before adaptation is shown for reference. Step sizes chosen on validation data. Ten seeds for SGAC, full precision and QLoRA with kept adapters, five for the other methods. The last column is the mean paired difference over the seeds the two methods share, so it can differ from the difference of the printed averages by the rounding of each to three decimals. $^*$paired $t$-test, $p<0.05$; in the bottom block every interval that excludes zero has $p<0.05$. $^\dagger$Not a 4/8-bit network: the full-precision adapters stay beside the codes at inference.
\end{tablenotes}
\end{threeparttable}
\end{adjustbox}
\end{table}

Table~\ref{tab:main} scores the network each method leaves on the device, and its bottom block averages the two cameras seed by seed. On this average, SGAC's deployed network has the highest mean accuracy of the nine low-bit networks. Its margin over the other eight ranges from 0.2 points, against the PEQA-style baseline, to 3.1 points against direct low-bit training, and every paired CI excludes zero (last column). The PEQA-style baseline comes closest: the 0.2-point difference [$+$0.1, $+$0.4] is detectable ($p\,{=}\,0.02$) but lies inside the equivalence margin (TOST $p\,{=}\,0.004$), so on average the two methods are equivalent. QLoRA with its full-precision adapters kept beside the codes is 0.3 points above SGAC and also equivalent to it (TOST $p\,{=}\,0.001$), but it is not a 4/8-bit network and it keeps 11.5 instead of 8.1 bits of state.

The two cameras separate the methods in different ways. On camera~A, the milder shift, SGAC and full precision quantized after training are equivalent (difference $-$0.1 points, CI [$-$0.5, $+$0.2]), and QAT, PEQA and merged QLoRA at ranks 1 and 8 fall 0.7 to 1.4 points behind ($p\le0.014$). Camera~B is the stronger shift, and here the return to codes matters most. Many updates of full precision and of QLoRA are smaller than a code step and do not survive it: merged QLoRA at rank 4 reaches 0.560 on the new camera against 0.618 with its adapters kept, and full precision falls to 0.593. SGAC writes its updates back as they accumulate and keeps 0.618, which puts it 1.1 to 2.4 points above full precision, QAT, ECO and merged QLoRA at ranks 4 and 8 (all $p<0.02$) and level with merged QLoRA at rank 1 (TOST $p\,{=}\,0.049$). The PEQA-style baseline re-tunes the per-group scales, which suits the global brightness and color change of camera~B; it is 0.5 points above SGAC on this camera (CI [$+$0.1, $+$0.8]) and 0.9 points below it on camera~A.

\textbf{The deployed format changes the ranking.} Scored with its adapters kept, QLoRA at rank 4 would be second of all methods on camera~B. Scored as the network the device runs, it is sixth of the nine methods that end in codes. Direct low-bit training, which rounds every weight at every step and keeps no remainder, stays 2.9 and 3.4 points below SGAC on the two cameras; the cache, and not the low-bit storage, carries the result. Fig.~\ref{fig:qual}(d)--(e) shows SGAC correcting the keypoints under both cameras. On one CPU thread an SGAC step takes 1.49\,s, close to QLoRA (1.35\,s) and faster than QAT (1.59\,s).

\textbf{How the score weights learning and forgetting.} The average in Table~\ref{tab:main} gives the new and the original images equal weight. Weighting the new camera by $\lambda$ and the original images by $1-\lambda$, as in (\ref{eq:jlambda}), and recomputing the ranking from the per-seed results, SGAC's deployed network ranks first among the nine low-bit networks for every $\lambda$ from 0.40 to 1. At 0.35 SGAC and the ECO-style baseline tie, and below that weight the ECO-style baseline, which keeps the most of the original images, ranks first, as expected when forgetting is nearly all that counts.

\begin{figure}[!tbp]
\centering
\includegraphics[width=0.62\textwidth]{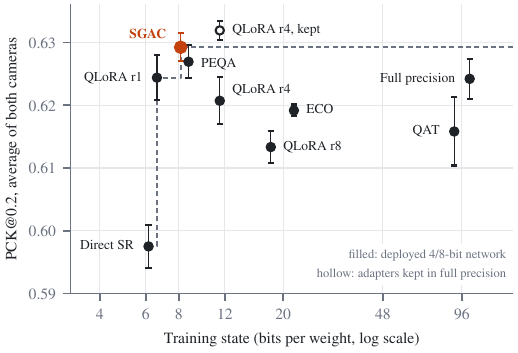}
\caption{Training state against the accuracy of the deployed network, averaged over both cameras (mean $\pm$ std over seeds). The dashed line is the frontier of the networks that end in 4/8-bit codes: the best accuracy reachable with at most that much state. Full precision and QAT both use 96 bits and are drawn slightly apart.}
\label{fig:pareto}
\end{figure}

\textbf{Memory and accuracy together.} Fig.~\ref{fig:pareto} places every method by its training state and its averaged accuracy. Among the networks that end in 4/8-bit codes, no method with more state than SGAC is more accurate, and the two methods with less state, merged QLoRA at rank 1 (6.6 bits) and direct low-bit training (6.1 bits), are 0.5 and 3.1 points below it. QLoRA with kept adapters lies above this frontier because it is not a 4/8-bit network. The same-storage rivals of Section~\ref{sec:rivals}, which include QA-LoRA at 5.7 bits, were run separately and are compared there.

\subsection{A Standard Corruption Benchmark: FreiHAND-C}
\label{sec:fhc}

\begin{table}[!tbp]
\caption{Deployed 4/8-bit PalmNet-K$_0$ after adaptation on FreiHAND-C, twelve ImageNet-C corruptions at severity 3 (PCK@0.2, mean over three seeds; $\pm$ std for mPC and average).}
\label{tab:fhc}
\centering
\footnotesize
\setlength{\tabcolsep}{3.5pt}
\begin{adjustbox}{max width=\textwidth}
\begin{threeparttable}
\begin{tabular}{@{}lccccccccc@{\hspace{4pt}}l@{}}
\toprule
& & \multicolumn{4}{c}{Category mean (no.\ of corruptions)} & & & & \multicolumn{2}{c}{SGAC$-$method (pts)} \\
\cmidrule(lr){3-6}\cmidrule(l){10-11}
Method & \begin{tabular}[c]{@{}c@{}}State\\(bits/w)\end{tabular} & Noise (3) & Blur (4) & Weather (2) & Digital (3) & \begin{tabular}[c]{@{}c@{}}mPC\\(12)\end{tabular} & Clean & Average & mPC [95\% CI] & average [95\% CI] \\
\midrule
Before adaptation, full precision & -- & 0.397 & 0.593 & 0.586 & 0.555 & 0.533 & 0.679 & 0.606 & -- & -- \\
\midrule
Full precision + Adam, then 4/8-bit & 96 & 0.458 & 0.608 & 0.608 & 0.588 & 0.565{\scriptsize$\pm$0.006} & 0.667 & 0.616{\scriptsize$\pm$0.004} & $+$0.8 [$-$0.3, $+$2.0] & $+$0.6 [$-$0.1, $+$1.3] \\
QAT, 32-bit master copy & 96 & \textbf{0.481} & 0.605 & 0.609 & \textbf{0.602} & \textbf{0.574}{\scriptsize$\pm$0.005} & 0.657 & 0.616{\scriptsize$\pm$0.005} & 0.0 [$-$0.8, $+$0.7] & $+$0.6 [$-$0.3, $+$1.5] \\
QLoRA rank 4, merged & 11.47 & 0.448 & 0.613 & 0.605 & 0.585 & 0.564{\scriptsize$\pm$0.004} & \textbf{0.678} & 0.621{\scriptsize$\pm$0.002} & $+$1.0 [$-$0.4, $+$2.4] & $+$0.1 [$-$0.7, $+$0.9] \\
PEQA-style (grid only) & 8.72 & 0.420 & 0.611 & \textbf{0.619} & 0.588 & 0.559{\scriptsize$\pm$0.005} & 0.676 & 0.617{\scriptsize$\pm$0.003} & $+$1.5 [$+$0.9, $+$2.0] & $+$0.5 [$+$0.1, $+$0.8] \\
ECO-style (momentum feedback) & 21.98 & 0.410 & 0.611 & 0.601 & 0.573 & 0.550{\scriptsize$\pm$0.001} & 0.674 & 0.612{\scriptsize$\pm$0.001} & $+$2.4 [$+$1.9, $+$3.0] & $+$1.0 [$+$0.3, $+$1.7] \\
Direct low-bit (SR) & 6.14 & 0.453 & 0.589 & 0.587 & 0.578 & 0.552{\scriptsize$\pm$0.007} & 0.633 & 0.593{\scriptsize$\pm$0.009} & $+$2.2 [$+$0.7, $+$3.6] & $+$2.9 [$+$1.1, $+$4.8] \\
\midrule
\textbf{SGAC (ours)} & \textbf{8.14} & 0.461 & \textbf{0.615} & 0.618 & \textbf{0.602} & \textbf{0.574}{\scriptsize$\pm$0.002} & 0.670 & \textbf{0.622}{\scriptsize$\pm$0.002} & -- & -- \\
\textit{QLoRA rank 4, adapters kept$^\dagger$} & 11.47 & 0.447 & 0.617 & 0.620 & 0.598 & 0.571{\scriptsize$\pm$0.004} & 0.672 & 0.621{\scriptsize$\pm$0.003} & $+$0.3 [$-$0.2, $+$0.8] & 0.0 [$-$0.5, $+$0.6] \\
\bottomrule
\end{tabular}
\begin{tablenotes}[flushleft]\footnotesize
\item[] Columns: mean PCK@0.2 over the corruptions of each ImageNet-C category; mPC, the mean over all twelve corruptions \cite{wang2021cococ}; Clean, the uncorrupted evaluation hands; Average, the mean of mPC and Clean. Every method keeps the step size selected for the cameras (Table~\ref{tab:main}), without re-tuning; differences are paired by seed. $^\dagger$Not a 4/8-bit network: the full-precision adapters stay beside the codes.
\end{tablenotes}
\end{threeparttable}
\end{adjustbox}
\end{table}

The two cameras are our own simulation. To add a standard benchmark, we follow the common-corruption protocol of ImageNet-C \cite{hendrycks2019}, which COCO-C applies to human pose estimation \cite{wang2021cococ} and which is the usual test bed for test-time adaptation \cite{tent}. We use twelve of its fifteen corruptions at severity~3: Gaussian, shot and impulse noise; defocus, glass, motion and zoom blur; fog and brightness; contrast, pixelation and JPEG compression. Frost and snow need external textures, and the elastic transform moves pixels and would invalidate the keypoint labels, so these three are omitted. Parameters with a spatial unit were defined for 224-pixel images and are scaled to the 96-pixel crop. Following the mixed-domain setting of test-time adaptation, the 12,000 adaptation crops carry the twelve corruptions in turn, and the 1,000 evaluation hands are tested under each corruption separately and on their clean originals. Every method keeps the step size selected for the cameras, without re-tuning.

Table~\ref{tab:fhc} reports the deployed networks over three seeds. We summarize robustness by the mean PCK@0.2 over the twelve corruptions (mPC), as COCO-C does \cite{wang2021cococ}, and add the clean evaluation hands to track forgetting. SGAC's deployed network has the highest average of corrupted and clean accuracy (0.622) and shares the highest mPC with QAT (0.574); QAT reaches it with a 96-bit master copy and is 1.3 points lower on the clean images. Against the PEQA-style, ECO-style and direct low-bit baselines, SGAC's lead holds on both measures with paired CIs that exclude zero, and it improves on direct low-bit training under every one of the twelve corruptions. Full precision quantized after training, QAT and merged QLoRA lie within 0.6 points of SGAC on the average, and with three seeds their intervals include zero. The categories differ. Under noise, QAT's master copy is 2.0 points ahead of SGAC; under blur and the digital corruptions SGAC is at or near the top; and under the weather corruptions the PEQA-style baseline, which re-tunes the grid, does as well as SGAC, as it did on the brightness and color change of camera~B. Merged QLoRA keeps the clean images best (0.678) but recovers less of the corrupted accuracy (0.564 mPC), which is the same trade-off we observed on the two cameras.

\subsection{The Closest Rivals on the Same Storage}
\label{sec:rivals}

\begin{table}[!tbp]
\caption{Same-storage rivals on camera A: deployed PCK@0.2 (mean $\pm$ std over five seeds) and paired differences.}
\label{tab:rivals}
\centering
\small
\setlength{\tabcolsep}{5pt}
\begin{adjustbox}{max width=\textwidth}
\begin{threeparttable}
\begin{tabular}{@{}lcccccc@{}}
\toprule
Method & \begin{tabular}[c]{@{}c@{}}State\\(bits/w)\end{tabular} & New camera & Original & Average & \begin{tabular}[c]{@{}c@{}}SGAC$-$method, new\\camera, pts [95\% CI]\end{tabular} & \begin{tabular}[c]{@{}c@{}}SGAC$-$method,\\average, pts [95\% CI]\end{tabular} \\
\midrule
Before adaptation, full precision & -- & 0.558 & 0.679 & 0.618 & -- & -- \\
\midrule
\textbf{SGAC (ours)} & 8.14 & \pmstd{\textbf{0.599}}{0.004} & \pmstd{0.640}{0.004} & \pmstd{0.620}{0.004} & -- & -- \\
QA-LoRA rank 4 \cite{qalora} & 5.70 & \pmstd{0.592}{0.005} & \pmstd{0.652}{0.009} & \pmstd{0.622}{0.006} & $+$0.6 [$+$0.3, $+$0.9] & $-$0.3 [$-$0.9, $+$0.4] \\
LoQT rank 4 \cite{loqt} & 10.94 & \pmstd{0.571}{0.002} & \pmstd{\textbf{0.675}}{0.002} & \pmstd{\textbf{0.623}}{0.001} & $+$2.8 [$+$2.2, $+$3.4] & $-$0.3 [$-$0.9, $+$0.2] \\
SpIEL, 1.5\% deltas \cite{spiel}, merged & 7.44 & \pmstd{0.574}{0.002} & \pmstd{0.672}{0.002} & \pmstd{0.623}{0.002} & $+$2.5 [$+$2.1, $+$2.9] & $-$0.3 [$-$0.7, $+$0.1] \\
\textit{SpIEL, 1.5\% deltas \cite{spiel}, kept$^\dagger$} & 7.44 & \pmstd{0.584}{0.003} & \pmstd{0.663}{0.003} & \pmstd{0.624}{0.003} & $+$1.5 [$+$1.2, $+$1.7] & $-$0.4 [$-$0.6, $-$0.2] \\
\bottomrule
\end{tabular}
\begin{tablenotes}[flushleft]\footnotesize
\item[] Separate CPU runs with their own data order, so this table is compared within, not with Table~\ref{tab:main}. Step sizes chosen on validation data. $^\dagger$Not a 4/8-bit network: the full-precision deltas stay beside the codes.
\end{tablenotes}
\end{threeparttable}
\end{adjustbox}
\end{table}

Table~\ref{tab:rivals} runs the three methods closest to SGAC on the same 4/8-bit storage, from the same checkpoint, with the same data, 400 steps and the same training of outliers and biases. QA-LoRA \cite{qalora} trains rank-4 adapters that change only the group offsets $\beta$, so its merge into the deployed network is exact. LoQT \cite{loqt} trains a rank-4 factor beside every frozen layer and merges it into the codes after 50, 150, 350 and 400 steps. SpIEL \cite{spiel} keeps full-precision deltas for 1.5\% of the weights, re-selects them every 20 steps by their gradients, and merges them into the codes at the end; we also report it with its deltas kept beside the codes. The step size of every method, SGAC included, is chosen on the validation data (three or four values each), and the chosen settings run with five seeds on camera~A.

Over five seeds, the two halves of the average tell the methods apart (Table~\ref{tab:rivals}). SGAC learns the new camera best, with paired CIs that exclude zero against all three rivals: it is 0.6 points [0.3, 0.9] ahead of QA-LoRA and 2.8 and 2.5 points ahead of LoQT and SpIEL. It moves the codes themselves wherever the cache admits a group, whereas each rival restricts the update: QA-LoRA to the group offsets, SpIEL to 1.5\% of the weights, and LoQT to a rank-4 subspace per layer. The same restriction lets the rivals keep more of the original images, and on the average of the two test sets the three deployed rivals lie 0.3 points above SGAC, with CIs that include zero in every case. SGAC reaches this with 8.1 bits of training state, between QA-LoRA (5.7) and LoQT (10.9), and it needs no merge step: the network on the device is the trained network after every step.

\subsection{Real Photographs and a Real Camera}
\label{sec:real}

\textbf{Longer adaptation and real photographs.} We repeated the keypoint comparison at full scale with the final PalmNet-K: 30,000 crops, 3,000 steps with batch size 128, the common step size and one seed. Here the starting codes are the 4/8-bit codes quantized after training, without quantization-aware training, so they start lower than the full-precision weights (0.470 on camera~A and 0.625 on original images, against 0.591 and 0.728), and every low-bit method must also recover the quantization loss. On camera~A, SGAC reaches 0.660 on the new camera and 0.675 on the original images, QLoRA with its adapters kept 0.665 and 0.661, and QAT with a 32-bit master copy 0.671 and 0.661. We then adapted the network toward real photographs: COCO-WholeBody hands photographed in the wild by many cameras and people \cite{cocowholebody}. They are part of its training data, yet they are the test set on which it is weakest, so this experiment measures targeted adaptation toward a weak real domain rather than adaptation to an unseen one. The network adapts on COCO-WholeBody training hands and is tested on its validation hands (target), FreiHAND and RHD (original). Before adaptation, its 4/8-bit codes reach 0.562, 0.628 and 0.537, against the full-precision scores of Table~\ref{tab:keypoints}. SGAC reaches 0.704, 0.716 and 0.713, QLoRA with kept adapters 0.710, 0.717 and 0.709, and QAT 0.715, 0.699 and 0.691. On real photographs, SGAC and QLoRA with kept adapters are within 0.6 points of each other on every test set, and SGAC is the most accurate 4/8-bit network on RHD.

\textbf{A real camera and an unseen user.} The guided calibration of our system (Appendix~\ref{app:system}) recorded one user that the networks had never seen through a webcam they had never seen, with PalmNet-K$_0$ as the network to adapt: ten prompted hand shapes, a calibration round and a held-out test round of 140 crops. The keypoint teacher, refined by our pose prior, labels the calibration crops, and 62 of 281 are kept, those that are consistent under a horizontal flip and agree with the prompted shape. Adapting the deployed 4/8-bit PalmNet-K$_0$ on them, mixed one-to-one with FreiHAND crops, raises the share of test crops read as the prompted shape from 0.20 to 0.42 with SGAC (mean of three seeds, range 0.39--0.44), to 0.41 with direct low-bit training and to 0.48 with merged QLoRA; every gain over the starting network is significant in every seed (exact McNemar $p\le0.003$ \cite{mcnemar}), while with 140 test crops the differences between the three methods are not (paired $t$-test over seeds, $p\ge0.37$). What separates them is forgetting: FreiHAND PCK@0.2 rises from 0.679 to 0.692 with SGAC and to 0.693 with merged QLoRA, but falls to 0.660 with direct low-bit training.

\subsection{Two Further Networks}
\label{sec:second}

\begin{table}[!tbp]
\caption{SGAC on two further networks, deployed at 4/8 bits: the hand finder PalmNet-D$_0$ on COCO-WholeBody frames (hand F1) and SoundTag with a new microphone (clip accuracy); means over five seeds.}
\label{tab:second}
\centering
\small
\setlength{\tabcolsep}{5pt}
\begin{adjustbox}{max width=\textwidth}
\begin{threeparttable}
\begin{tabular}{@{}lccccc@{}}
\toprule
& \multicolumn{3}{c}{Finder (F1)} & \multicolumn{2}{c}{SoundTag (accuracy)} \\
\cmidrule(lr){2-4}\cmidrule(l){5-6}
Method & COCO (target) & HaGRID & RHD & New mic & Original \\
\midrule
Before adaptation, full precision & 0.422 & 0.764 & 0.775 & -- & -- \\
\midrule
Full precision, then 4/8-bit & 0.469 & 0.737 & 0.776 & 0.629 & 0.436 \\
QLoRA rank 4, merged & 0.445 & 0.787 & 0.796 & 0.567 & 0.488 \\
PEQA-style (grid only) & 0.466 & 0.737 & 0.785 & 0.619 & 0.420 \\
ECO-style & 0.455 & 0.738 & 0.767 & 0.635 & 0.499 \\
Biases only ($B=0$)$^a$ & 0.463 & 0.747 & 0.784 & -- & -- \\
Direct low-bit (SR) & 0.389 & 0.566 & 0.641 & 0.591 & 0.372 \\
\textbf{SGAC (ours)} & \textbf{0.471} & 0.740 & 0.790 & \textbf{0.638} & 0.423 \\
\textit{QLoRA, adapters kept}$^\dagger$ & 0.470 & 0.748 & 0.791 & 0.635 & 0.428 \\
\bottomrule
\end{tabular}
\begin{tablenotes}[flushleft]\footnotesize
\item[] $^a$Four seeds. $^\dagger$Not a 4/8-bit network: the full-precision adapters stay beside the codes at inference.
\end{tablenotes}
\end{threeparttable}
\end{adjustbox}
\end{table}

Table~\ref{tab:second} repeats the comparison on two networks with different structure and data, with the same protocol and five seeds each. The hand finder PalmNet-D$_0$ (157k parameters) adapts to in-the-wild COCO-WholeBody frames \cite{cocowholebody}. These frames are among its training sources, together with HaGRID \cite{hagrid}, RHD and composites, but before adaptation it finds hands in them with a full-precision F1 of only 0.422, against 0.764 on HaGRID and 0.775 on RHD. SGAC raises the COCO F1 to 0.471 while keeping 0.740 on HaGRID and 0.790 on RHD. Averaging the target domain with the mean of the two original domains, as for the keypoint network, SGAC is ahead of full precision quantized after training by 0.5 points ($p\,{=}\,0.02$), of the ECO-style baseline by 1.4 points ($p<0.001$) and of biases-only training by 0.4 points ($p\,{=}\,0.007$, four seeds); the PEQA-style baseline is not distinguishable ($+$0.5, $p\,{=}\,0.32$), and direct low-bit training loses 13 to 20 points on the original domains. Merged QLoRA reaches the same average (0.618) by another route: re-quantization rounds away most of its update, so it keeps the original domains but retains only 0.445 on COCO, against 0.470 with its adapters kept.

SoundTag, a 218k-parameter audio tagger trained from scratch on ESC-50 \cite{esc50}, adapts to a new microphone with a narrow band and a fan-noise floor. The shift is strong enough that every method trades original accuracy for the new microphone. SGAC reaches the highest new-microphone accuracy, 0.638, and 0.423 on the original recordings; full precision, merged QLoRA, the PEQA-style baseline and QLoRA with kept adapters are all within 1.1 points of SGAC on the average ($p\ge0.17$). The ECO-style baseline, whose momentum lets updates that never reach a code step decay, keeps more of the original recordings (0.499) at 22 instead of 8.1 bits of state.

\subsection{Ablation Study}
\label{sec:ablation}

\begin{table}[!tbp]
\caption{Ablation of SGAC on camera A at a common step size (PCK@0.2, mean $\pm$ std over three seeds).}
\label{tab:ablation}
\centering
\small
\setlength{\tabcolsep}{2.0pt}
\begin{adjustbox}{max width=\textwidth}
\begin{threeparttable}
\begin{tabular}{@{}lccccc@{}}
\toprule
Configuration & \begin{tabular}[c]{@{}c@{}}State\\(bits/w)\end{tabular} & New camera & Original & Average & $\Delta$avg (pts) \\
\midrule
SGAC ($B=2$, four ways, $\pi_g$) & 8.14 & \pmstd{0.586}{0.001} & \pmstd{0.667}{0.003} & \pmstd{0.627}{0.002} & -- \\
\midrule
\multicolumn{6}{@{}l}{\textit{One component changed}} \\
admission: random & 8.14 & \pmstd{0.523}{0.012} & \pmstd{0.624}{0.010} & \pmstd{0.574}{0.011} & $-$5.3 \\
admission: Fisher only ($n_gF_g$) & 8.14 & \pmstd{0.588}{0.001} & \pmstd{0.668}{0.001} & \pmstd{0.628}{0.001} & $+$0.2 \\
plain gradient instead of (\ref{eq:compass}) & 8.14 & \pmstd{0.584}{0.004} & \pmstd{0.669}{0.004} & \pmstd{0.627}{0.003} & $+$0.0 \\
uncached groups rounded, not waiting & 8.14 & \pmstd{0.576}{0.010} & \pmstd{0.648}{0.006} & \pmstd{0.612}{0.003} & $-$1.4 \\
no cache ($B=0$) & 6.14 & \pmstd{0.564}{0.001} & \pmstd{0.676}{0.000} & \pmstd{0.620}{0.000} & $-$0.7 \\
direct-mapped cache (one way) & 8.14 & \pmstd{0.587}{0.003} & \pmstd{0.669}{0.001} & \pmstd{0.628}{0.002} & $+$0.1 \\
fully associative cache & 8.14 & \pmstd{0.586}{0.006} & \pmstd{0.668}{0.004} & \pmstd{0.627}{0.005} & $+$0.0 \\
\midrule
\multicolumn{6}{@{}l}{\textit{Cache budget}} \\
$B=1$ (12\% of groups) & 7.14 & \pmstd{0.579}{0.002} & \pmstd{0.667}{0.003} & \pmstd{0.623}{0.002} & $-$0.4 \\
$B=4$ (47\% of groups) & 10.14 & \pmstd{0.590}{0.004} & \pmstd{0.667}{0.002} & \pmstd{0.629}{0.002} & $+$0.2 \\
\midrule
\multicolumn{6}{@{}l}{\textit{Residual state for every weight, same codes and update}} \\
8-bit accumulator, every weight$^a$ & 14.64 & \pmstd{0.592}{0.002} & \pmstd{0.664}{0.001} & \pmstd{0.628}{0.001} & $+$0.1 \\
4-bit residual, every weight$^b$ & 10.64 & \pmstd{0.594}{0.005} & \pmstd{0.650}{0.002} & \pmstd{0.622}{0.001} & $-$0.5 \\
2-bit residual, every weight$^b$ & 8.64 & \pmstd{0.574}{0.012} & \pmstd{0.647}{0.011} & \pmstd{0.611}{0.004} & $-$1.6 \\
\bottomrule
\end{tabular}
\begin{tablenotes}[flushleft]\footnotesize
\item[] $^a$The arrangement of in-memory mixed-precision training \cite{nandakumar2020}, an accumulator for every weight with whole-step transfer, here in 8 bits. $^b$The arrangement of Error Diffusion \cite{errordiffusion}, a low-bit residual for every weight, quantized by stochastic rounding.
\end{tablenotes}
\end{threeparttable}
\end{adjustbox}
\end{table}

Table~\ref{tab:ablation} changes one component of SGAC per row, on camera~A at the common step size of $1.5\times10^{-4}$, with three seeds. Both halves of wait-or-write-back matter.

\textbf{Admission.} With random admission the average falls by 5.3 points ($p\,{=}\,0.013$), below the network before adaptation: the lines go to groups whose updates matter little while the sensitive groups wait. Fisher-only admission ($n_gF_g$) performs on par with $\pi_g$ ($+$0.2 points, $p\,{=}\,0.19$). On this network the code steps are similar across groups, and the two scores pick about three quarters of the same groups (Section~\ref{sec:checks}); on a regression whose groups differ in curvature and code step, $\pi_g$ beats Fisher-only admission in 9 of 10 seeds, which is why we keep the step factor.

\textbf{Waiting.} Rounding the uncached groups at every step instead of letting them wait costs 1.4 points ($p\,{=}\,0.031$), most of it on the original images ($-$1.9 points against $-$1.0 on the new camera). This is what Proposition~\ref{prop:wait} predicts: rounding noise raises the loss on data the network already knows. Without a cache ($B=0$), only outliers and biases learn; the original images are kept, but the new camera gains 2.2 points less ($p\,{=}\,0.002$). The plain gradient, a direct-mapped and a fully associative cache all stay within 0.2 points of the default.

\textbf{Cache budget and residuals for every weight.} The average rises from 0.620 without a cache to 0.623 at $B=1$ and 0.627 at $B=2$, and it barely moves beyond that, so about a quarter of the groups carries the useful updates. The bottom block of Table~\ref{tab:ablation} keeps residual state for every weight instead, with the same codes and the same update. An 8-bit accumulator for every weight, the arrangement of in-memory mixed-precision training \cite{nandakumar2020}, reaches 0.628 with 14.6 bits of state: the accuracy of SGAC with 80\% more state, so SGAC needs 56\% of it. A 2-bit residual for every weight, the arrangement of Error Diffusion \cite{errordiffusion} at roughly SGAC's memory, reaches 0.611, 1.6 points lower, and a 4-bit residual needs 10.6 bits to come within 0.5 points. At equal memory, a large buffer for the most sensitive groups is worth more than a small buffer for all of them. Restricting full-precision updates to the 23\% of weights with the largest Fisher information \cite{fishmask}, or pulling them toward the starting weights with an $L_2$ penalty, does not protect the original images as well: both keep at most 0.654 on them, against 0.667 for SGAC on the same seeds.

\subsection{Checks of the Analysis}
\label{sec:checks}

\textbf{Unbiasedness and the exact rounding cost.} Over 400 trials with a forced eviction, the mean move of each of 64 codes equals the exact accumulated update within sampling error (max $|z|=3.3$), as Proposition~\ref{prop:unbiased} states. On a quadratic with 256 weights and updates of up to 0.9 code steps, one rounding step adds a measured loss of 0.114 ($\pm$0.004), against 0.112 from the exact expression (\ref{eq:first}) and 0.153 from the bound $\tfrac18 s^2\sum H$. For steps of at most 0.05 code steps it adds 0.0158, which (\ref{eq:first}) predicts exactly, while the bound stays at 0.153 and the exact step gains only 0.0086. Small steps make the bound loose and the rounding cost larger than the gain, as Proposition~\ref{prop:first} predicts.

\begin{table}[!tbp]
\caption{Wait or round, measured on the 6,163 weight groups of the deployed PalmNet-K$_0$ on camera A (gradients over 48 batches; curvature from 48 Hutchinson probes, summed per group).}
\label{tab:waitround}
\centering
\small
\setlength{\tabcolsep}{3pt}
\begin{adjustbox}{max width=\textwidth}
\begin{threeparttable}
\begin{tabular}{@{}lcc@{}}
\toprule
Quantity & \begin{tabular}[c]{@{}c@{}}Start of\\adaptation\end{tabular} & \begin{tabular}[c]{@{}c@{}}After 400\\SGAC steps\end{tabular} \\
\midrule
\multicolumn{3}{@{}l}{\textit{Groups for which waiting beats rounding, (\ref{eq:wait})}} \\
$\lambda=\tfrac12$ (new and original images) & 5.6\% & \textbf{36.6\%} \\
\ \ with 12 / 24 probes & 6.6 / 5.8\% & 37.0 / 36.7\% \\
$\lambda=0$ (original images only) & 25.7\% & 43.9\% \\
$\lambda=1$ (new camera only) & 1.6\% & 19.2\% \\
same decision with 24 and 48 probes & 98.1\% & 94.3\% \\
per-weight curvature clamped at 0, 12 / 48 probes$^a$ & 23.3 / 13.9\% & 65.6 / 55.5\% \\
\midrule
\multicolumn{3}{@{}l}{\textit{Original images}} \\
first-order gain / rounding cost & $+$13.6 & $-$1.65 \\
\midrule
\multicolumn{3}{@{}l}{\textit{Line value of Corollary~\ref{cor:admission} held by the top 23\% of groups}} \\
ranked by $\pi_g$ & 76.0\% & 69.0\% \\
ranked by Fisher only ($n_gF_g$) & 77.9\% & 71.4\% \\
ranked by $V_g$ itself (best possible) & 86.1\% & 89.7\% \\
groups shared by $\pi_g$ and Fisher-only ranking & 76.7\% & 75.6\% \\
\bottomrule
\end{tabular}
\begin{tablenotes}[flushleft]\footnotesize
\item[] $^a$Clamping each noisy diagonal estimate before summing biases the curvature term upward when probes are few; the group sums above are unbiased.
\end{tablenotes}
\end{threeparttable}
\end{adjustbox}
\end{table}

\textbf{Waiting and rounding on PalmNet-K.} Table~\ref{tab:waitround} evaluates Proposition~\ref{prop:wait} on all 6,163 weight groups of the deployed keypoint network on camera~A, once at the start of adaptation and once after 400 SGAC steps. Gradients are averaged over 48 batches (bias-corrected). The curvature term $\sum_i H_{ii}\,\mathbb{E}|g_i|$ of each group is estimated from 48 Hutchinson probes \cite{hutchinson}, summing the raw diagonal estimates over the group before any clipping, which keeps the group term unbiased; the resulting wait-or-round decisions agree for 94--98\% of the groups between 24 and 48 probes. At the start, the network has just met the new camera and its expected gradient is strong and consistent: rounding beats waiting for 94\% of the groups, and the update even helps the original images, whose first-order gain is 13.6 times their rounding cost. After 400 steps the update has turned against the original images ($-1.65$ times their rounding cost), and waiting is now the better choice for 37\% of the groups at $\lambda=\tfrac12$, for 44\% when only the original images count and for 19\% when only the new camera counts. The direction of the shift is the same for every $\lambda$. Summed over the groups outside the cache, the one-step model still favors rounding at both points; the 400-step ablation of Section~\ref{sec:ablation} measures the net effect of a whole adaptation directly, and there rounding the uncached groups costs 1.4 points, most of them on the original images.

\textbf{How good is the admission score.} The ideal line value $V_g$ of Corollary~\ref{cor:admission} needs $G$ and $H$, which the device does not track. With the estimates of Table~\ref{tab:waitround}, the 23\% of groups with the largest $\pi_g$ hold 76\% of the total line value at the start of adaptation and 69\% after 400 steps, against 86\% and 90\% for the ideal ranking, so $\pi_g$ reaches 88\% and 77\% of the best possible cache content. On this network the Fisher-only score $n_gF_g$ captures as much (78\% and 71\%) and shares about three quarters of its groups with $\pi_g$, which agrees with the ablation, where the two rules are within 0.2 points. The step factor matters when code steps differ between groups: on a regression whose groups differ in curvature and code step, admission by $\pi_g$ reaches a loss of 8.83, against 9.45 for Fisher-only admission, 10.46 for random admission and 11.42 without a cache, and it beats random and Fisher-only admission in 8 and 9 of 10 seeds.

\subsection{Memory Traffic and Bit Allocation}
\label{sec:alloc}

\begin{table}[!tbp]
\caption{Training-state traffic per optimizer step (bits read and written per weight, counted from each update rule).}
\label{tab:traffic}
\centering
\small
\setlength{\tabcolsep}{3pt}
\begin{adjustbox}{max width=\textwidth}
\begin{tabular}{@{}lcc@{}}
\toprule
Method & Bits/weight/step & vs.\ full precision \\
\midrule
Full precision + Adam & 192 & 1 \\
QAT, 32-bit master copy & 196 & 1.02 \\
ECO-style (16-bit momentum) & 42.7 & 0.22 \\
QLoRA, rank 4 & 14.9 & 0.078 \\
Dense 2-bit residual & 14.7 & 0.077 \\
Direct low-bit (SR) & 10.7 & 0.056 \\
PEQA-style (grid only) & 7.0 & 0.036 \\
\textbf{SGAC} & \textbf{6.6} & \textbf{0.034} \\
\bottomrule
\end{tabular}
\end{adjustbox}
\end{table}

\textbf{Memory traffic.} The forward and backward passes are the same for every method; what differs is the training state that each optimizer step reads and writes. Table~\ref{tab:traffic} counts it from the update rules with the numbers of Table~\ref{tab:bits}. A full-precision weight with Adam moves 192 bits per step. SGAC reads and writes one 32-bit Fisher average per group, the 8-bit accumulators of the cached 23\% of groups, and a code only when it commits, which happens $2.3\times10^{-4}$ times per weight and step. It therefore moves 6.6 bits per weight per step, 29 times less than full precision and 2.3 times less than QLoRA at rank 4, whose optimizer state is small but touched at every step. On small devices an off-chip memory access costs roughly two orders of magnitude more energy per bit than an arithmetic operation \cite{horowitz}, so state traffic is a reasonable proxy for the energy of the optimizer step.

\textbf{Bit allocation.} The deployed layout keeps the first layer and the head at 8 bits and the rest at 4 bits. We checked it against a layer-wise allocation that ranks layers by Fisher-weighted quantization error, raises the most sensitive ones to 8 bits and keeps a few outlier channels, at a budget of about 5 bits, with 12 minutes of quantization-aware training per network. Post-training quantization to uniform 4 bits costs 20 points of PCK@0.2 for the final PalmNet-K (0.728 in full precision to 0.531 on FreiHAND) and 22 points of F1 for PalmNet-D$_0$ (0.76 to 0.54 on HaGRID), so 4-bit codes need training. Once trained, uniform 4-bit codes already match full precision: 0.736 for PalmNet-K at 192\,KB, and an F1 of 0.78 for PalmNet-D$_0$ at 79\,KB, against 0.76 at 631\,KB in full precision. The layer-wise allocation reaches the same accuracy (0.736 at 5.0 bits; 0.78 at 5.5 bits) with 25--38\% more memory and helps only on the cross-dataset RHD hands (0.814 against 0.807 for PalmNet-D$_0$). We therefore keep the 8/4-bit layout, so the codes that SGAC trains are the codes that are deployed.

\FloatBarrier
\section{The Host System}
\label{sec:system}

SGAC was built for a two-hand webcam interface (Fig.~\ref{fig:system}). The right hand directs a person and the left hand a cat; a hand-and-face finder (PalmNet-D) locates both hands in every frame, PalmNet-K reads 21 joints per hand, a causal gesture network turns each keypoint stream into gesture tokens, and an intent grammar maps the tokens to actions given the state of the scene. All networks are trained from scratch, stored in the 4/8-bit format of Section~\ref{sec:storage} and exported to JavaScript and WebGL, so the system runs in a browser; together they have about 0.8M parameters. A two-minute guided calibration collects labeled crops from the user's own camera, and these crops are what SGAC adapts on.

Appendix~\ref{app:system} describes the components, and Appendix~\ref{app:results} reports their results. In brief: the causal gesture network lowers false alarms on generated free-motion streams from 115 per minute for per-frame finger rules to 0.3, and to zero with a three-frame hold; a diffusion prior of hand poses, guided by the keypoint network's own heatmaps, adds 4.4 to 4.7 points of PCK@0.2 at test time without changing a weight; the same prior, conditioned on the finger states that a gesture class fixes, turns gesture-labelled hands into 21-joint labels; and a 1.1M-parameter keypoint network trained with these labels on upright crops is ahead of MediaPipe Hands \cite{mediapipe}, scored on every hand of the test set, on COCO-WholeBody and RHD.

\FloatBarrier
\section{Conclusion}

SGAC trains a low-bit network in place by wait-or-write-back. The codes are the only copy of the convolution weights; a small cache of 8-bit accumulators, admitted by a sensitivity score and flushed without bias, carries the updates that are smaller than one code step, and every other group waits instead of being rounded. Our analysis shows when waiting helps: rounding a sub-step update costs loss at the same order as the update itself, and once the network has adapted, the update turns against the data it already knew; over a full adaptation, letting the uncached groups wait is worth 1.4 points. Scored on the network the device actually runs, SGAC's deployed keypoint network had the highest mean accuracy of the nine low-bit networks we compared on two simulated cameras and the highest average of corrupted and clean accuracy on FreiHAND-C; it was on par with the best baselines on real photographs and on a hand finder, and it learned the new microphone best on a sound network. It keeps 8.1 bits of training state per weight, against 11.5 for QLoRA at rank 4 and 96 for a 32-bit master copy with Adam, and moves 2.3 and 29 times less state per step than these two. Because the network that learns is the network that runs, nothing has to be converted after adaptation. By (\ref{eq:memory}), the 157k-weight finder needs 161\,KB of training state with SGAC against 1.9\,MB with a 32-bit master copy and Adam, and the 383k-weight keypoint network needs 390\,KB against 4.6\,MB. The training state of either network therefore fits in the 512\,KB of on-chip SRAM of a microcontroller such as the ESP32-S3, where the master-copy state of neither does.

\FloatBarrier
\section*{Acknowledgments}
The hand images of Appendix~\ref{app:results} were produced with a text-to-image generative model, and the hand-shape label of every image was verified manually. 

\clearpage
\appendix

\section{Host System Components}
\label{app:system}

\begin{algorithm}[!tbp]
\caption{One frame of the two-hand system}
\label{alg:frame}
\footnotesize
\begin{algorithmic}[1]
\Require frame $I_t$, audio window, previous keypoints $P_{t-1}^h$, feature buffers
\State Locate both hands and the face with PalmNet-D; give each hand its role by its side of the mirrored frame
\State Update the motion $\mu_h$ of each hand by (\ref{eq:motion}) and choose its level $v_h$ under the stagger rule (\ref{eq:stagger})
\For{$h\in\{R,L\}$}
 \If{$v_h=\textit{reuse}$} \State $P_t^h\gets$ decoder and head of PalmNet-K on the buffered features
 \Else \State $P_t^h\gets$ PalmNet-K on the crop of hand $h$; store its encoder features \EndIf
 \State Refine $P_t^h$ with the guided pose prior
 \State $p^h\gets$ GestureTCN on the last 32 frames; add the gesture memory by (\ref{eq:mem})
 \State Emit $y_t^h=c$ if $p^h_c\ge\tau_c$ of (\ref{eq:fdr}) for three frames
\EndFor
\State Tag the audio window with SoundTag
\State \Return actions $\gets\mathcal{G}(y_t^R,y_t^L,\text{tags},s_t)$
\end{algorithmic}
\end{algorithm}

Algorithm~\ref{alg:frame} summarizes one frame. Single-image hand pose estimation regresses keypoint score maps from a hand crop \cite{zimmermann2017}; multiview bootstrapping produced labels at scale \cite{simon2017}, and transformer models trained on millions of hands reach a few millimeters of error \cite{hamer,wilor}, while mobile models trade accuracy for speed \cite{mobrecon,mediapipe}. Our components follow the lightweight end of this range.

\textbf{Finding the hands (PalmNet-D).} A $96\times128$ frame is mapped to a $12\times16$ grid with three maps: where hand centers are, how large each palm is (the logarithm of the palm length), and where the face is (157k parameters). Each hand peak gives a hand crop, each hand takes its role from its side of the mirrored frame, and the face is used for expressions and head motion. Trained on rendered hands composited into real rooms (Places365 \cite{places}) and on labelled real frames from HaGRID \cite{hagrid}, COCO-WholeBody and RHD, the final finder reaches a full-precision hand F1 of 0.773 on HaGRID test users it has never seen (0.816 at $192\times256$), against 0.30 when trained on rendered frames only.

\textbf{Reading 21 joints (PalmNet-K).} A small U-Net (Fig.~\ref{fig:qual}(f)) with encoder widths 24, 48, 64 and 96 and 383k parameters reads a $96\times96$ crop around each hand and predicts 21 keypoint heatmaps, 21 depth maps and a hand-presence score. It is trained with a spatial-softmax cross-entropy loss on the heatmaps; the keypoints are decoded from the heatmap peaks with sub-pixel refinement, and the joints that decide a gesture, such as the finger tips and the thumb--index interface, receive a larger weight in the loss. Both networks are built from \textbf{CompassConv} blocks, which train six branches with batch normalization ($3\times3$, $1\times3$, $3\times1$, $1\times1$, an axial ``$+$'' and a diagonal ``$\times$'' stencil) and fold them exactly into one $3\times3$ convolution for inference, following structural re-parameterization \cite{acnet,dbb}.

\textbf{Pose prior.} A heatmap can be wrong in a way no other joint can see: a fingertip with two peaks, or a joint that jumps to the wrong finger. We therefore refine each skeleton with a learned prior of hand shapes. Diffusion models learn a data distribution by denoising \cite{sohl2015,ddpm}, which is score matching \cite{vincent2011}; DDIM samples them deterministically in few steps \cite{ddim}, and classifier guidance adds the gradient of a classifier to steer the samples \cite{dhariwal2021}. Diffusion priors of body pose have been used as regularizers \cite{dposer} and guided by 2D detections to fit 3D bodies \cite{scorehmr}. We apply the same combination to 2D hand skeletons. A denoising diffusion model on the 42 coordinates of a skeleton, a 0.5M-parameter MLP whose time step enters every residual block as a scale and shift, learns how plausible hands look. The decoded skeleton is noised to a moderate level and denoised back with ten deterministic DDIM steps, and at every step the predicted clean skeleton is pulled up the log-likelihood of PalmNet-K's own heatmaps, with the heatmaps in the role of the classifier. The likelihood keeps every mode of a heatmap: for each joint it is a mixture of small Gaussians at the 64 most probable heatmap cells, weighted by their probability. The prior is trained on FreiHAND training poses that share no pose with our validation and adaptation data, and it changes no weight of PalmNet-K.

\textbf{Gestures in a continuous stream.} Gestures arrive in a stream with no marked start or end, and most of the time no gesture is performed at all \cite{kopuklu2019}; a system that treats every movement as a command raises false alarms, the Midas touch problem \cite{jacob1990}. Online gesture recognition has used recurrent 3D convolutional networks \cite{molchanov2016}, light detectors that gate a heavy classifier \cite{kopuklu2019} and skeleton networks small enough for a CPU \cite{ddnet}, and false positives per hour of everyday motion have been used to measure false alarms \cite{magic}. For each gesture class $c$, our detector should find as many gestures as possible while the false discovery rate stays below $\alpha$:
\begin{equation}
\max_{\tau_c}\ \mathrm{D}_c(\tau_c)\ \ \text{s.t.}\ \ \mathrm{FDR}_c(\tau_c)=\frac{\mathrm{FP}_c(\tau_c)}{\max(1,\mathrm{D}_c(\tau_c))}\le\alpha,
\label{eq:p2}
\end{equation}
where $\tau_c$ is a threshold on the class probability, and $\mathrm{D}_c$ and $\mathrm{FP}_c$ count all detections and the false ones. A causal temporal convolutional network (GestureTCN) \cite{tcn} with kernel size 5, dilations 1, 2 and 4, and 48 channels (47k parameters) reads the last 32 frames of one hand, about one second at 30\,fps: its keypoints, normalized for position, scale and rotation, together with the keypoint velocities, palm facing, pinch distance and finger extensions. It is pre-trained by masked finger modeling, in which a finger, a time span or the thumb--index interface is hidden and reconstructed, and then trained to name the gesture from the third frame of the gesture on, so gestures are named early. A gesture fires when its probability stays above $\tau_c$ for three frames, and a sticky hidden Markov model (HMM) smooths the probabilities over time. Because the number of detections $\mathrm{D}_c(\tau)$ can only fall as $\tau$ grows, the best threshold that meets the constraint of (\ref{eq:p2}) is the smallest feasible one \cite{bh1995}:
\begin{equation}
\tau_c=\min\big\{\tau:\ \mathrm{FDR}_c(\tau)\le\alpha\big\},\qquad \alpha=1\%,
\label{eq:fdr}
\end{equation}
counted over the frames of 24 calibration streams on a grid from 0.30 to 0.99. In the interactive demonstration, the network output is also multiplied by rule-based finger states (a product of experts) as a safeguard for real webcam keypoints.

\textbf{Turning gestures into actions.} The intent grammar $\mathcal{G}(\cdot,s_t)$ gives each token its meaning given the scene state $s_t$. A rule names a token, a hand, a precondition on the scene, a hold time and a cooldown. The right hand (the person) has six rules: an open palm stops everyone, a fist bends the person down to the cat, a thumb up stands her back up, two bent fingers pet the cat while she is bent down and wave hello while she stands, and an index toward the cat says ``Sit!''. The left hand (the cat) has four: a fist makes the cat sit, an open palm makes it ignore you, a flip of the hand makes it swat, and a thumb up makes it lick your hand or your shoe. Because two bent fingers mean ``pet'' only while the person is bent down near the cat, the grammar removes a class of false actions that no per-frame classifier can remove, and it runs inside the real-time loop. SoundTag (a four-block CNN with attention pooling over a log-mel spectrogram, 218k parameters) tags 50 sound classes, so the characters react to a knock or a clap.

\textbf{Guided calibration.} On-device adaptation needs labeled frames from the user's own camera. In a two-minute guided calibration, the user shows ten prompted right-hand poses (open palm, back of the hand, fist, pointing up and sideways, V sign upright and tilted, two fingers walking, pinch and a relaxed hand); PalmNet-D tracks the hand, and a square crop around it is stored five times a second, in a calibration round and a shorter test round that is kept apart. All frames stay on the device. The prompt labels the hand shape of every crop, and keypoint pseudo-labels come from the keypoint teacher refined by the pose prior, kept only when they are consistent under a horizontal flip and agree with the prompted shape. These frames are the labeled samples of (\ref{eq:p3}).

\textbf{Run-time memory for two hands.} A hand that does not move does not need a full network pass; for video, sparse convolutions reuse the features of unchanged content \cite{deltacnn}. For each hand we keep an exponential average of its keypoint motion, normalized by its palm size $r_h$,
\begin{equation}
\mu_h\leftarrow\lambda_\mu\mu_h+(1-\lambda_\mu)\,\tfrac{1}{21}\textstyle\sum_{j}\|p_j^h(t)-p_j^h(t-1)\|/r_h,
\label{eq:motion}
\end{equation}
and below a threshold the hand \emph{reuses} its encoder features from a temporal buffer (at most four frames old), so only the decoder and the head run. A stagger rule allows at most one hand below full depth on each frame,
\begin{equation}
\mathbf{1}[v_R(t)\neq\mathrm{full}]\cdot\mathbf{1}[v_L(t)\neq\mathrm{full}]=0,
\label{eq:stagger}
\end{equation}
where $v_h(t)$ is the level of hand $h$. A gesture memory per role stores prototype hand shapes with 8/4-bit keys. A query retrieves its top-8 keys with one step of a modern Hopfield network \cite{hopfield}, and their class distribution $p_{\mathrm{mem}}$ corrects the classifier output $p$ when the classifier is uncertain:
\begin{equation}
\log\tilde p=\log p+\gamma\,\mathbf{1}\big[\mathcal{H}(p)>h_0\big]\log p_{\mathrm{mem}},
\label{eq:mem}
\end{equation}
where $\mathcal{H}$ is the entropy, and $\gamma$ and $h_0$ are chosen on validation streams.

\FloatBarrier
\section{Results of the Host System}
\label{app:results}

\begin{table}[!tbp]
\caption{Gesture recognition on held-out generated streams, one change per row (mean $\pm$ std over streams).}
\label{tab:gesture}
\centering
\small
\setlength{\tabcolsep}{2.2pt}
\begin{adjustbox}{max width=\textwidth}
\begin{tabular}{@{}lcccc@{}}
\toprule
Configuration & \begin{tabular}[c]{@{}c@{}}False alarms\\/min (free)\end{tabular} & \begin{tabular}[c]{@{}c@{}}Wrong calls\\/min\end{tabular} & \begin{tabular}[c]{@{}c@{}}Detected\\(\%)\end{tabular} & \begin{tabular}[c]{@{}c@{}}Latency from\\onset (ms)\end{tabular} \\
\midrule
Finger rules, per frame & \pmstd{115.2}{24.8} & \pmstd{71.0}{24.7} & \pmstd{97.0}{4.0} & \pmstd{12}{13} \\
\ \ + 3-frame hold & \pmstd{43.4}{6.9} & \pmstd{30.3}{8.3} & \pmstd{96.9}{4.0} & \pmstd{28}{24} \\
\midrule
GestureTCN, per frame & \pmstd{0.3}{0.5} & \pmstd{0.33}{0.60} & \pmstd{99.4}{1.9} & \pmstd{50}{19} \\
\ \ + 3-frame hold & \pmstd{0.0}{0.0} & \pmstd{0.20}{0.48} & \pmstd{99.3}{2.0} & \pmstd{105}{20} \\
\ \ + FDR thresholds (\ref{eq:fdr}) & \pmstd{0.0}{0.0} & \pmstd{0.00}{0.00} & \pmstd{96.4}{3.5} & \pmstd{223}{38} \\
\ \ + HMM & \pmstd{0.0}{0.0} & \pmstd{0.07}{0.25} & \pmstd{99.3}{2.0} & \pmstd{194}{18} \\
\ \ + rule product (demo) & \pmstd{1.3}{1.1} & \pmstd{1.67}{1.37} & \pmstd{98.4}{2.8} & \pmstd{151}{23} \\
\bottomrule
\end{tabular}
\end{adjustbox}
\end{table}

\begin{figure}[!tbp]
\centering
\includegraphics[width=0.6\textwidth]{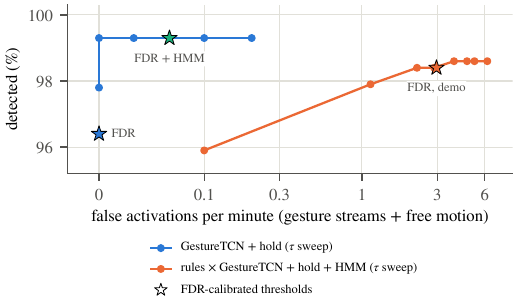}
\caption{Detection vs.\ false activations as the gesture threshold varies; stars mark the FDR-calibrated thresholds.}
\label{fig:gestcurve}
\end{figure}

\textbf{Gestures in a continuous stream.} The gesture streams are generated by a kinematic hand model: 160 training, 24 calibration and 30 test streams of 60\,s, plus 10\,min of free motion with hard negatives such as half-closed fists and passing fingers. Gestures are scored by the detection rate, the latency from the first frame of a gesture (which includes the time the hand takes to form it), the wrong calls per minute inside gesture streams, and the false alarms per minute of free motion, following the false-positives-per-hour measure of \cite{magic}. Table~\ref{tab:gesture} changes one component per row; every row uses the same trained GestureTCN. Per-frame finger rules fire 115 times per minute of free motion, and a three-frame hold still leaves 43; GestureTCN alone fires 0.3 times per minute and detects 99.4\% of the gestures, and the hold removes the remaining free-motion alarms. The FDR thresholds of (\ref{eq:fdr}) remove the wrong calls inside gesture streams but delay detection, and the HMM restores the detection rate (99.3\%) with 0.07 wrong calls per minute. In the demonstration pipeline, a fixed threshold of 0.5 instead of (\ref{eq:fdr}) raises free-motion alarms from 1.3 to $3.4\pm1.4$ per minute. Fig.~\ref{fig:gestcurve} shows the trade-off as the threshold varies. The per-frame network answers within 50\,ms of the first frame, and the zero-alarm settings trade 100--150\,ms for certainty, which suits discrete commands.

\begin{table}[!tbp]
\caption{Run-time memory for two hands.}
\label{tab:runtime}
\centering
\small
\setlength{\tabcolsep}{3pt}
\begin{adjustbox}{max width=\textwidth}
\begin{tabular}{@{}lcc@{}}
\toprule
\multicolumn{3}{@{}l}{(a) Motion gate, two 60-s hands $\times$ 4 streams, deployed PalmNet-K} \\
\midrule
Gating & Cost per frame & Frames with both hands reduced \\
\midrule
Always full depth & 1.00 & -- \\
Independent per hand & 0.76 & 17\% \\
\textbf{Staggered (ours)} & 0.80 & \textbf{0\%} \\
\midrule
\multicolumn{3}{@{}l}{(b) Gesture memory, 10,800 test frames per condition (accuracy)} \\
\midrule
Keypoint noise & Classifier & + memory, 8/4-bit keys \\
\midrule
none & 0.927 & \textbf{0.935} \\
2\% of palm size & 0.902 & \textbf{0.928} \\
5\% of palm size & 0.830 & \textbf{0.885} \\
\bottomrule
\end{tabular}
\end{adjustbox}
\end{table}

\textbf{Run-time memory for two hands.} Table~\ref{tab:runtime}(a) shows the motion gate of (\ref{eq:motion}) on gesture streams, with costs measured for the deployed PalmNet-K on one CPU thread, where a reuse pass costs 0.41 of a full pass. Gating each hand independently saves 24\% of the cost but reduces both hands at once in 17\% of the frames; the staggered gate saves 20\% and keeps one hand at full depth on every frame, as (\ref{eq:stagger}) guarantees, and the reused keypoints stay within 5.0\% of the palm length on average (8.2\% at the 95th percentile). Table~\ref{tab:runtime}(b) shows that the gesture memory of (\ref{eq:mem}) helps most for noisy keypoints (0.830 to 0.885). Keys at 8/4 bits match 32-bit keys and take 157\,KB instead of 691\,KB for 5,400 prototypes; 8--10\% of the queries change their top-1 key under quantization, but the full-precision top-1 key stays among the quantized top 8 for over 99.9\% of those queries, so the soft top-8 retrieval keeps every accuracy unchanged.

\begin{table}[!tbp]
\caption{The final PalmNet-K in full precision on four hand datasets.}
\label{tab:keypoints}
\centering
\small
\setlength{\tabcolsep}{4pt}
\begin{adjustbox}{max width=\textwidth}
\begin{tabular}{@{}lccc@{}}
\toprule
Test set & PCK@0.1 & PCK@0.2 & EPE (px) \\
\midrule
FreiHAND evaluation & 0.528 & 0.728 & 9.8 \\
RHD & 0.482 & 0.694 & 10.6 \\
COCO-WholeBody hands & 0.433 & 0.661 & 11.3 \\
CMU hands & 0.421 & 0.650 & 11.9 \\
\bottomrule
\end{tabular}
\end{adjustbox}
\end{table}

\begin{table}[!tbp]
\caption{Our networks, in full precision, and MediaPipe Hands on the same test images (PCK@0.2; hand finder: F1).}
\label{tab:mediapipe}
\centering
\small
\setlength{\tabcolsep}{3pt}
\begin{adjustbox}{max width=\textwidth}
\begin{tabular}{@{}lccccc@{}}
\toprule
& \multicolumn{2}{c}{MediaPipe Hands \cite{mediapipe}} & \multicolumn{3}{c}{Ours (from scratch)} \\
\cmidrule(lr){2-3}\cmidrule(l){4-6}
& & & PalmNet-K & \multicolumn{2}{c}{PalmNet-K, wide} \\
\cmidrule(l){5-6}
Keypoints & every hand & hands it finds & & two passes & tracked crop \\
\midrule
FreiHAND evaluation & 0.786 & 0.910 (86\%) & 0.728 & 0.771 & \textbf{0.845} \\
CMU hands & \textbf{0.703} & 0.873 (81\%) & 0.651 & -- & -- \\
COCO-WholeBody hands & 0.554 & 0.868 (64\%) & 0.661 & 0.692 & \textbf{0.795} \\
RHD & 0.601 & 0.870 (69\%) & 0.694 & 0.753 & \textbf{0.873} \\
Parameters & \multicolumn{2}{c}{2.0M (landmarks) $+$ 1.8M (palms)} & 383k & \multicolumn{2}{c}{1.1M} \\
\midrule
Hand finder & \multicolumn{2}{c}{MediaPipe palm detector} & PalmNet-D & \multicolumn{2}{c}{PalmNet-D, $192\times256$} \\
\midrule
HaGRID test users & \multicolumn{2}{c}{\textbf{0.933}} & 0.773 & \multicolumn{2}{c}{0.816} \\
RHD & \multicolumn{2}{c}{\textbf{0.862}} & 0.787 & \multicolumn{2}{c}{0.837} \\
COCO-WholeBody frames & \multicolumn{2}{c}{\textbf{0.550}} & 0.432 & \multicolumn{2}{c}{0.534} \\
User's webcam, hand in guide box & \multicolumn{2}{c}{0.969} & \textbf{0.979} & \multicolumn{2}{c}{0.929} \\
\bottomrule
\end{tabular}
\end{adjustbox}
\end{table}

\textbf{Keypoints and sound.} Table~\ref{tab:keypoints} reports the final PalmNet-K in full precision: PCK@0.2 stays between 0.65 and 0.73 across studio, rendered and in-the-wild hands, at 0.23\,MB and 5.7\,ms per crop on one CPU thread. PalmNet-K targets an accuracy--memory point rather than absolute accuracy. On the same FreiHAND images, MediaPipe Hands, which we ran ourselves, reaches PCK@0.2 of 0.91 on the 86\% of hands its detector finds, with a 7.6\,MB bundle \cite{mediapipe}, and transformer models with hundreds of millions of parameters are more accurate still \cite{hamer,wilor}. SoundTag reaches $70.9\pm2.2\%$ over the five ESC-50 folds \cite{esc50}, 6.4 points above the CNN of Piczak \cite{piczak2015}.

\textbf{Against MediaPipe Hands.} Table~\ref{tab:mediapipe} scores MediaPipe Hands, the public pretrained tracker, on the same test images in two ways: on every hand of a test set, where a hand its palm detector misses counts as wrong, and only on the hands it finds. Scored on every hand, the 383k-parameter PalmNet-K is ahead of MediaPipe on COCO-WholeBody (0.661 against 0.554) and RHD (0.694 against 0.601) and within six points on FreiHAND and CMU. MediaPipe's advantage lies in the hands it chooses to read: on those, its landmark network, trained on 16k manually labelled real photographs and 100k rendered hands \cite{mediapipe}, reaches PCK@0.2 of 0.87--0.91, and its palm detector, with eleven times the parameters of PalmNet-D, finds hands more reliably on photographs. The wide PalmNet-K described below narrows this gap: on a tracked crop it is ahead of MediaPipe's every-hand score on all three test sets, and on RHD it matches the accuracy MediaPipe reaches on only the hands it accepts (0.873 against 0.870), with 1.1M instead of 2.0M landmark parameters. On the user's webcam the two finders are level: the hand lies in the guide box in 97.9\% of the frames with PalmNet-D and 96.9\% with MediaPipe.

\begin{table}[!tbp]
\caption{Test-time refinement of the deployed 4/8-bit PalmNet-K$_0$ by a hand-pose prior (1,000 FreiHAND evaluation hands; no weight changes; every setting chosen on validation hands).}
\label{tab:prior}
\centering
\small
\setlength{\tabcolsep}{2.0pt}
\begin{adjustbox}{max width=\textwidth}
\begin{tabular}{@{}lcccccc@{}}
\toprule
& \multicolumn{2}{c}{Original} & \multicolumn{2}{c}{Camera A} & \multicolumn{2}{c}{Camera B} \\
\cmidrule(lr){2-3}\cmidrule(lr){4-5}\cmidrule(l){6-7}
Skeleton from & @0.1 & @0.2 & @0.1 & @0.2 & @0.1 & @0.2 \\
\midrule
Heatmap decode (argmax $+$ local mean) & 0.459 & 0.680 & 0.325 & 0.545 & 0.231 & 0.430 \\
Multi-mode heatmap likelihood, no prior & 0.473 & 0.685 & 0.336 & 0.551 & 0.239 & 0.435 \\
$+$ Gaussian (PCA) prior, MAP & 0.482 & 0.704 & 0.339 & 0.570 & 0.241 & 0.457 \\
$+$ VAE prior, MAP & 0.495 & 0.714 & \textbf{0.352} & 0.582 & \textbf{0.251} & 0.469 \\
$+$ diffusion prior, MAP & \textbf{0.497} & 0.716 & 0.352 & 0.579 & 0.248 & 0.464 \\
$+$ diffusion prior, guided DDIM (ours) & 0.488 & \textbf{0.726} & 0.342 & \textbf{0.589} & 0.242 & \textbf{0.477} \\
\midrule
\multicolumn{7}{@{}l}{\textit{Crops shifted, scaled and rotated at random (detector-like crops)}} \\
Heatmap decode & 0.456 & 0.671 & 0.307 & 0.525 & 0.229 & 0.428 \\
$+$ diffusion prior, guided DDIM & \textbf{0.490} & \textbf{0.709} & \textbf{0.337} & \textbf{0.570} & \textbf{0.252} & \textbf{0.467} \\
\bottomrule
\end{tabular}
\end{adjustbox}
\end{table}

\textbf{The pose prior.} Table~\ref{tab:prior} refines the skeletons of the deployed 4/8-bit PalmNet-K$_0$, before any adaptation, with the pose prior and two simpler priors; nothing in the network changes. Every setting (noise level, number of steps, guidance strength, prior weight and likelihood width) is chosen on 1,000 validation hands, by the mean PCK over the three conditions, and the table reports the evaluation hands. The guided diffusion prior raises PCK@0.2 by 4.6 points on original images (0.680 to 0.726), 4.4 points through camera~A (0.545 to 0.589) and 4.7 points through camera~B (0.430 to 0.477), and lowers the end-point error by about 8\%; on PCK@0.1 its paired gain is 2.9 points (95\% bootstrap CI [2.3, 3.6]) on original images. Re-reading the heatmaps with the multi-mode likelihood alone gains only 0.5--0.6 points, so the gain comes from the prior. A Gaussian prior over the 42 coordinates gains 2.4--2.7 points, and learned priors gain more: a VAE \cite{vae} and the diffusion prior used for gradient-based MAP refinement are within one point of each other, and the guided DDIM sampler is best at PCK@0.2 on all three conditions, while MAP refinement is best at the stricter PCK@0.1. The gain does not depend on crops centered on the true hand: when every test crop is shifted by up to 10\% of its size, scaled by 0.8--1.25 and rotated by up to 30$^\circ$, as a detector would deliver it, the guided prior still adds 3.8--4.5 points. Refining one skeleton takes ten passes of the 0.5M-parameter prior, 18\,ms on one CPU thread.

\textbf{Keypoint labels from gesture classes.} A gesture label fixes which fingers are extended, and nothing else. We condition the pose prior on these finger states (folded, extended or free for each finger) and train it with classifier-free guidance \cite{cfg}: each finger state is hidden with probability 0.15 and the whole condition with probability 0.15, so one network is both conditional and unconditional. A reading of the teacher is refined as above with the noise prediction $\hat\epsilon=\epsilon_\varnothing+w\,(\epsilon_c-\epsilon_\varnothing)$. On 1,280 fully labelled COCO-WholeBody validation hands, with the finger states taken from the labels as a gesture class would give them, the teacher alone places 0.692 of the joints within PCK@0.2 and reads every finger state right on 24.5\% of the hands; the unconditional prior ($w=0$) gives 0.734 and 29.4\%, and the conditional prior gives 0.746 and 42.8\% at $w=3$, 0.750 and 46.0\% at $w=5$, and 0.748 and 41.9\% at $w=7$, while stronger guidance overshoots (0.719 and 20.6\% at $w=15$). Guidance at $w=5$ nearly doubles the share of hands whose shape is read right and also moves the joints closer to the truth, so every gesture-labelled hand can become a 21-joint training label.

\textbf{Upright crops and gesture-labelled hands.} MediaPipe reads every hand in an upright crop: the hand is rotated so that the line from the wrist to the middle knuckles points up, and a square twice the size of the palm box is cut around it \cite{mediapipe}. We train a wider PalmNet-K, with encoder widths 40, 80, 112 and 160, 1.1M parameters and a $128\times128$ input, on such crops with random rotations of up to $20^\circ$, distilled from the teacher after the teacher has been fine-tuned on the same crops. Two kinds of labels come from HaGRID \cite{hagrid}, whose hands carry only a gesture class. A finger-state loss on over 100,000 of its training hands asks the network's soft keypoints to keep each finger extended or folded as the class says, and the conditional prior at $w=5$ refines the teacher's reading of 19,734 HaGRID hands: the refined finger states agree with the gesture class on 14,000 of them instead of 11,169, and the 12,313 hands whose palm stays in place and whose fingers agree become 21-joint labels. With FreiHAND, RHD, COCO-WholeBody and HInt \cite{hamer}, the student is trained on 244k crops. Read in two passes (a first reading places the upright crop for the second), it reaches PCK@0.2 of 0.771 on FreiHAND, 0.692 on COCO-WholeBody and 0.753 on RHD, against 0.728, 0.661 and 0.694 for the 383k network, which puts it 13.8 and 15.2 points ahead of MediaPipe's every-hand score on COCO-WholeBody and RHD. On a tracked crop, placed from the hand's joints as a tracker places it from the previous frame, it reaches 0.845, 0.795 and 0.873, and 0.792 on the HInt test hands. On the 109 frames of the user's webcam recording whose hand fills the crop tightly, it reads 0.61 of the prompted hand shapes correctly in two passes, against 0.30 for the final 383k PalmNet-K in full precision, without any calibration (the 4/8-bit PalmNet-K$_0$ of Section~\ref{sec:real} reads 0.20 of the 140 test-round crops before adaptation); MediaPipe, trained on manually labelled real photographs, reads 0.88, and the guided calibration is the route by which the wide network adapts to the user on the device.

\textbf{Generated hand images.} Hand shapes such as a pointing finger or a V at an unusual tilt are rare in the keypoint datasets. We generated two sets of photographs with an image model: 369 hand-centered tiles of nine hand shapes and five kinds of hard negatives, and 324 webcam scenes at arm's length with pointing and V hands at many tilts and with two hands. The hand-shape label of every image was verified manually, so every test is scored against human-verified labels. The images carry no keypoint ground truth: the keypoint teacher provides pseudo-labels on the training batches, kept only when they are consistent under a horizontal flip and when the hand shape they imply agrees with the human label (154 of 647 crops); every fourth generation batch is held out for testing. Before this filter the teacher's hand shape agrees with the human label on 40.8\% of the crops (Cohen's $\kappa=0.31$ \cite{cohen1960}), so the filter, not the teacher, makes the pseudo-labels reliable. Fine-tuning PalmNet-K on the pseudo-labels, mixed one-to-one with FreiHAND crops, is compared with the same fine-tuning on FreiHAND alone, paired by seed, over three seeds. The share of hand shapes read correctly rises from 0.30 to 0.49 on the held-out tiles (exact McNemar $p\le0.003$ in every seed \cite{mcnemar}) and from 0.13 to 0.23 on the held-out scenes. On 589 real HaGRID test hands, photographed by other people with other cameras, it rises from 0.257 to 0.305 ($+$3.9 to $+$5.3 points in every seed, $p\le0.035$), while FreiHAND accuracy stays within 0.6 points of the control.

\end{document}